\documentclass[conference]{IEEEtran}

\IEEEoverridecommandlockouts                              % This command is only needed if 
\usepackage{times}
\usepackage[numbers]{natbib}
\usepackage{graphics} % for pdf, bitmapped graphics files
\usepackage{mathtools}
\usepackage{graphicx}
\usepackage{tabularx}
\usepackage{caption}
\usepackage{subcaption}
\usepackage{wrapfig}
\usepackage{romannum}
\usepackage{amsmath,amssymb}
\usepackage{url}
\usepackage{bm}
\usepackage[table]{xcolor}
\usepackage[pagebackref=true,breaklinks=true,colorlinks=true,bookmarks=false,citecolor=blue]{hyperref} % make magenta color refs and urls, blue color cites
\usepackage{multirow}
\usepackage{booktabs}
\usepackage{pifont}

\usepackage[ruled,vlined]{algorithm2e}
\usepackage[noend]{algpseudocode}
\makeatletter
\let\OldStatex\Statex
\renewcommand{\Statex}[1][3]{%
  \setlength\@tempdima{\algorithmicindent}%
  \OldStatex\hskip\dimexpr#1\@tempdima\relax}
\renewcommand{\ALG@beginalgorithmic}{\normalsize}

\newcommand{\vb}{{\bf b}}

\newcommand{\vx}{{\bf x}}

\newcommand{\vz}{{\bf z}}

\newcommand{\vo}{{\bf o}}

\newcommand{\vu}{{\bf u}}

\newcommand{\vW}{{\bf W}}

\newcommand{\vR}{{\bf R}}
\newcommand{\vE}{{\bf E}}

\newcommand{\vF}{{\bf F}}

\newcommand{\VSigma}{{\mbox{\boldmath$\Sigma$}}}

\newcommand{\vmu}{{\mbox{\boldmath$\mu$}}}

\newcommand{\calL}{\mathcal{L}}
\newcommand{\calD}{\mathcal{D}}

\newcommand{\calN}{\mathcal{N}}

\begin{document}

\title{Bridging Active Exploration and Uncertainty-Aware Deployment Using Probabilistic Ensemble Neural Network Dynamics}

\author{Taekyung Kim$^*$, Jungwi Mun$^*$, Junwon Seo, Beomsu Kim, and Seongil Hong \\
AI Autonomy Technology Center, Agency for Defense Development
\thanks{$^{*}$These authors contributed equally to this work}%
}

\makeatletter
\let\@oldmaketitle\@maketitle% Store \@maketitle
\renewcommand{\@maketitle}{\@oldmaketitle% Update \@maketitle to insert...
  \includegraphics[width=0.98\linewidth]{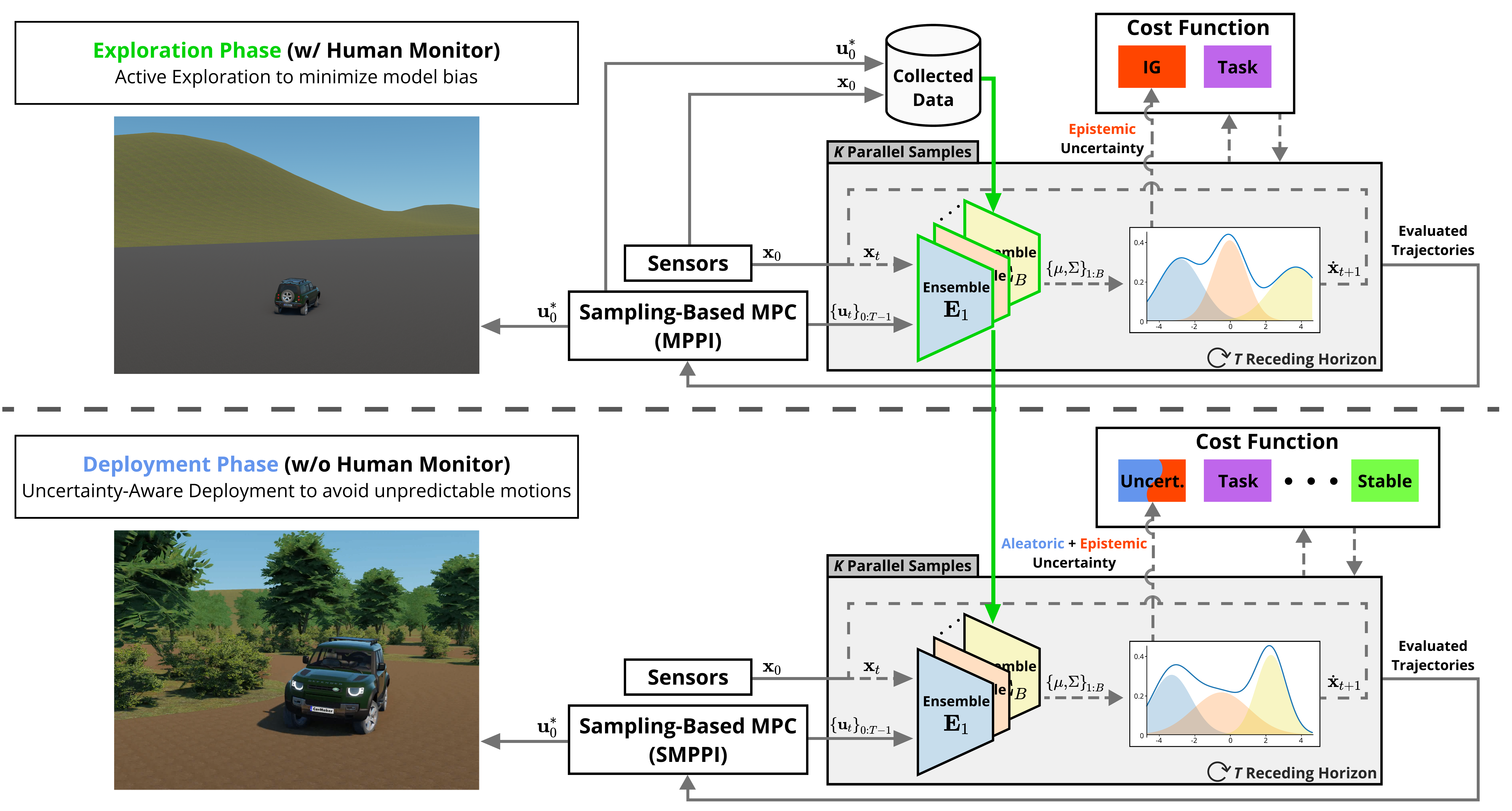}
  \captionof{figure}{Overview diagram of our unified model-based reinforcement learning framework with dynamics learning. In \textit{exploration phase}, a parallelized ensemble neural network serves as the robot dynamics and outputs the estimated posterior distribution of the next state. To enable active exploration, we quantify epistemic uncertainty by measuring the ensemble disagreement via Jensen-Rényi Divergence. In \textit{deployment phase}, the neural network dynamics trained during the active exploration phase is applied directly to perform uncertainty-aware control. We transfer the neural network dynamics for uncertainty-aware deployment with minimal modification. Project page: \url{https://taekyung.me/rss2023-bridging}
  }
  \label{fig:main}
\vspace{-0.05in}
  \bigskip}
\makeatother

\maketitle
\addtocounter{figure}{-1} % decrease 1 on figure count (the second figure is displayed from 3 to 2)
%%%%%%%%%%%%%%%%%%%%%%%%%%%%%%%%%%%%%%%%%%%%%%%%%%%%%%%%%%%%%%%%%%%%%%%%%%%%%%%%

\begin{abstract}
In recent years, learning-based control in robotics has gained significant attention due to its capability to address complex tasks in real-world environments. With the advances in machine learning algorithms and computational capabilities, this approach is becoming increasingly important for solving challenging control problems in robotics by learning unknown or partially known robot dynamics. Active exploration, in which a robot directs itself to states that yield the highest information gain, is essential for efficient data collection and minimizing human supervision. Similarly, uncertainty-aware deployment has been a growing concern in robotic control, as uncertain actions informed by the learned model can lead to unstable motions or failure. However, active exploration and uncertainty-aware deployment have been studied independently, and there is limited literature that seamlessly integrates them. This paper presents a unified model-based reinforcement learning framework that bridges these two tasks in the robotics control domain. Our framework uses a probabilistic ensemble neural network for dynamics learning, allowing the quantification of epistemic uncertainty via Jensen-Rényi Divergence. The two opposing tasks of exploration and deployment are optimized through state-of-the-art sampling-based MPC, resulting in efficient collection of training data and successful avoidance of uncertain state-action spaces. We conduct experiments on both autonomous vehicles and wheeled robots, showing promising results for both exploration and deployment.
\end{abstract}
\IEEEpeerreviewmaketitle

%%%%%%%%%%%%%%%%%%%%%%%%%%%%%%%%%%%%%%%%%%%%%%%%%%%%%%%%%%%%%%%%%%%%%%%%%%%%%%%%

\section{Introduction}

Learning-based robotic control has been an increasingly active research topic in recent decades, pushing the boundary of robotic control tasks with new capabilities through model-free reinforcement learning~\cite{morimoto_acquisition_2001, peng_deeploco_2017, hwangbo_learning_2019, kumar_rma_2021}, model-based reinforcement learning~\cite{abbeel_application_2007, deisenroth_pilco_2011, englert_model-based_2013, williams_information_2017, kahn_uncertainty-aware_2017, nagabandi_dexterous_2019, wu_daydreamer_2022}, and robot dynamics learning~\cite{nguyen-tuong_using_2010, kabzan_learning-based_2019, lutter_deep_2019, oconnell_neural-fly_2022}. There are two key aspects that are commonly taken into account while deploying an intelligent robot. The first is to assure safety, and the second, if at all possible, is to achieve maximum performance.

Model-based methods have been regarded as being advantageous in robotic applications over pure model-free strategies for various reasons. Firstly, sample efficiency is essential since real-world samples are highly expensive in terms of time, labor, and finances~\cite{kober_reinforcement_2013}. Secondly, humans are primarily more intuitive about how to incorporate prior knowledge into a model compared to a policy or value function~\cite{lutter_deep_2019, kim_physics_2022}. Lastly, models are task-agnostic and may thus be utilized to optimize arbitrary cost functions, whereas the majority of model-free policies are bounded to a specific task. 

These benefits have led to recent control paradigms that include dynamics learning into a model-based control framework, outperforming the conventional methods that rely on manually tuned models based on human insights into physics~\cite{nagabandi_dexterous_2019, williams_information-theoretic_2018, yang_data_2019}. This scheme may also be described as a model-based reinforcement learning problem, as the model is continuously trained to provide a more accurate representation of the system, while an optimization-based controller serves as the policy. Such a control framework possesses the inherent flexibility to dynamically adjust during execution, for instance, to changing target speeds or actuator limits. Gaussian Process~(GP) is a typical regression method in model learning problems due to its advantageous features, such as the ability to cope with small data samples and incorporate prior knowledge~\cite{deisenroth_pilco_2011, williams_gaussian_1995}. However, it does not scale well to large datasets and to high-dimensional systems. To mitigate this issue, recent robotics studies have benefited from neural networks owing to their universal function approximation properties with high model capacity. 

The limitation of these model-based control frameworks is that the accuracy of the learned model drastically affects the control performance~\cite{nguyen-tuong_model_2011}. However, the real-world is fraught with uncertainty, making the model learning problem complex and challenging. There are two primary viewpoints within the robotics community that tackle inevitable uncertainty in opposing ways. 

A spontaneous solution would be to prevent the robot from entering uncertain state-action spaces to evade unpredictable motions. This is a straightforward strategy, given that an inaccurate prediction of the dynamics model might result in substantial losses in control performance and, in the worst case, instability. We refer to such an uncertainty-averse strategy as \textit{uncertainty-aware deployment} (or safety-aware deployment), and it is usually applied during execution after completing the robot's training.

% \begin{figure*}[t]
% \centering
% \includegraphics[width=0.99\textwidth]{Figures/Main.pdf}
% \caption{Overview diagram of our unified model-based reinforcement learning framework with dynamics learning.}
% \label{fig:main}
% \vspace*{-0.05in}
% \end{figure*}

The exact opposite of the above strategy is to deliberately visit unexplored state-action spaces that provide high uncertainty. The hypothesis is that by trying to explore uncertain segments of the learned model, it can learn more precise representation of the system with fewer samples and discover behaviors with higher performance. Such uncertainty-seeking strategy is known as \textit{active exploration}, and it is considered during the early phase of training data exploration. 

We argue that these opposite yet analogous strategies addressing uncertainty must be entailed concurrently in a learning-based control framework. No matter how much data is gathered through active exploration, the uncertainty that eventually has not been eliminated needs to be monitored and avoided throughout robot deployment. Conversely, when we take the model uncertainties for risk-averse planning, they are only meaningful if the model is trained with the right data; otherwise, the controller will produce erroneous motions. Nevertheless, these two approaches have largely been studied independently, and there is little literature that integrates them in a seamless manner. 

This paper introduces a unified model-based reinforcement learning framework in a robot dynamics learning setting, seamlessly bridging active exploration and uncertainty-aware deployment to fulfill both safety and maximum performance. Fig.~\ref{fig:main} shows that the proposed framework consists of two phases. In \textit{exploration phase}, a parallelized ensemble neural network serves as the robot dynamics and outputs the estimated posterior distribution of the next state. In \textit{deployment phase}, the neural network dynamics trained during the active exploration phase is applied directly to perform uncertainty-aware control. Both tasks are optimized using the state-of-the-art sampling-based Model Predictive Contorl (MPC), which, owing to its property, allows the insertion of arbitrary cost functions after training.

In summary, the contributions of this paper are as follows:
\begin{itemize}
    \item We introduce a fully parallelized probabilistic ensemble neural network that is sensitive to uncertainty in order to learn robot dynamics.
    \item We separate epistemic uncertainty from model disagreement for active exploration.
    \item We transfer the neural network dynamics for uncertainty-aware deployment with minimal modification.
    \item We conduct extensive experiments with our framework on both autonomous vehicles and wheeled robots, outperforming the compared methods by a large margin.
\end{itemize}

\section{Related Works}
\subsection{Model Learning with Uncertainty}

A regression model capable of accurately capturing the nonlinear dynamics is essential for model-based control in uncertain environments with nonlinearity. Data-driven models approximating the robot's forward dynamics have been extensively studied in numerous prior research, including GP~\cite{deisenroth_pilco_2011, williams_gaussian_1995}, time-varying linear models~\cite{levine_learning_2014}, set-membership identification~\cite{milanese_optimal_1991, parsi_active_2020}, locally weighted projection regression~\cite{vijayakumar_incremental_2005, williams_locally_2019}, etc. GP has long been favored in dynamics learning given its simplicity and its inherent nature to account for both types of uncertainty: aleatoric uncertainty and epistemic uncertainty.

Aleatoric uncertainty arises from intrinsic stochasticities present in a system, such as observation noise and process noise. Epistemic uncertainty, however, originates from a lack of sufficient data to uniquely determine the underlying characteristics of the target system in an approximated model. An ideal model trained with an infinite number of data may have zero epistemic uncertainty, but it is not feasible to eradicate it completely in a realistic setting. 

Neural networks are known for being tractable to large training datasets while assuring constant-time inference and having the potential to learn highly complex and even non-smooth robot dynamics. The recent breakthrough in the field of dynamics learning is that neural networks may also capture both types of uncertainty by using an ensemble of probabilistic neural networks~\cite{chua_deep_2018}. Vlastelica et al.~\cite{vlastelica_risk-averse_2021} proposed to quantify epistemic uncertainty by taking empirical variances over the means and variances of the propagated posterior distribution of each ensemble. We investigate further to determine a more exact epistemic uncertainty in order to provide a better exploration bonus (which is discussed in Section~\ref{sec:exp_active}). 

\subsection{Active Exploration}

Human-supervised data collection is time-consuming and can be biased by the fact that manual control of the robot may not encompass all of its possible motions. As a remedy, it was suggested that the agent should plan exploration by directing itself to the states that yield the largest information gain, therefore performing active exploration~\cite{thrun_efficient_1992}. In general, the measured model uncertainty is leveraged as the exploration motivation to minimize the required amount of data to gather~\cite{martinez-cantin_body_2010}.

We argue that the exploration strategy should be \textit{imaginary} and \textit{online}. If the agent is \textit{reactive} to immediate curiosity~\cite{pathak_curiosity-driven_2017}, which is the opposite to being \textit{imaginary}, it disregards the long-term effects of control actions to the system. Consequently, it must struggle in realms of underactuated continuous control~\cite{schultheis_receding_2019}. In addition, since most robotic systems are intrinsically stochastic, the real trajectories would quickly diverge from the planned ones. Therefore, it is desirable to plan future trajectory \textit{online} at every time step, rather than relying on the optimized trajectory found \textit{offline} as in~\cite{schultheis_receding_2019}.  

Several studies address model-based active exploration based on the above hypothesis. Examples of this line of research include the work in~\cite{shyam_model-based_2019, sekar_planning_2020, bucher_adversarial_2021}, in which model-free policies are employed to maximize expected information gain in a specific task. In~\cite{bechtle_curious_2019}, GP is used for model learning, while a gradient-based MPC handles the information gain acquired from the GP model. In this work, we adopt sampling-based MPC as a policy and demonstrate that it is scalable to handle uncertainty derived from neural networks. 

\subsection{Safe Deployment}

Uncertainty-averse deployment has been addressed as a major concern throughout robotic control history, and there is a growing interest in its application to model-based reinforcement learning. The uncertain actions informed by the learned model may result in unstable motions or even catastrophic failure. 

Kahn et al.~\cite{kahn_uncertainty-aware_2017} incorporated the estimated uncertainty as a cost to learn collision avoidance. Yu et al.~\cite{yu_mopo_2020} penalized uncertainty of the dynamics in an offline reinforcement learning setting to resolve the distributional shift issue. Wang et al.~\cite{wang_rough_2021} treated trajectories having greater uncertainty than a certain threshold as constraint conditions in an application of rough terrain navigation. Similar to~\cite{wang_rough_2021}, we penalize uncertainty with both soft and hard costs to maximize the control performance while ensuring safety.

\section{Methods}
\subsection{Probabilistic Ensemble Neural Network}
Consider a general continuous time system dynamics $\dot{\vx}_{t+1} = \vF(\vx_t, \vu_t)$ where $\vF$ is a nonlinear function, $\vx_t \in \mathbb{R}^{n}$ and $\vu_t \in \mathbb{R}^{m}$ are the observed state vector and applied action input at time~$t$, respectively. To account for both aleatoric and epistemic uncertainty, we use Probabilistic Ensemble Neural Network~(PENN) to approximate the unknown dynamics~$\vF$ via data-driven manner~\cite{chua_deep_2018, buckman_sample-efficient_2018}.

Each neural network predictive model~$\vE_{b}$ of the ensemble members outputs a Gaussian distribution conditioned on the model input:
\begin{equation}
    \vE_{b}(\vx_t, \vu_t; \theta_{b}) = \calN(\vmu_{\theta_{b}}(\vx_t, \vu_t), \VSigma_{\theta_{b}}(\vx_t, \vu_t))  \,,
    \label{eq:gaussian}
\end{equation}
where $\vmu$ and $\VSigma$ are the mean and the diagonal covariance parameterized by the model parameter~$\theta_{b}$. Given a collected dataset of transitions~$\calD \triangleq \{ (\vx_t, \vu_t, \dot{\vx}_{t+1})\}^{N-1}_{t=0}$ with the size of $N$, the probabilistic model can be trained with the Gaussian Negative Log-Likelihood~(NLL) loss function:
\begin{equation}
\begin{aligned}
    \calL_{\text{train}}(\theta_{b}) = \sum_{t=0}^{N-1}\left[\vmu_{\theta_{b}}-\dot{\vx}_{t+1}\right]^{\top} &  \VSigma_{\theta_{b}}^{-1} \left[\vmu_{\theta_{b}}-\dot{\vx}_{t+1}\right] \\
    & \quad + \log \left[ \operatorname{det} \VSigma_{\theta_{b}} \right]\,.
    \label{eq:train_loss}
\end{aligned}
\end{equation}
Note that random initialization is applied individually to each ensemble member, resulting in mutually independent initial weights. Consequently, the only difference between the ensemble members is their initial model parameters.

Consequently, the stochastic dynamics function predicting the next state can be interpreted as a Gaussian Mixture Model~(GMM) of the $B$ ensemble members:
\begin{equation}
    \vF(\theta_{1:B}) = \sum_{b=1}^{B}{\pi_b \vE_{b}(\vx_t, \vu_t; \theta_{b})} \,, \quad 0\leq \pi_{b} \leq 1 \,.
    \label{eq:penn}
\end{equation}
Here, we simply use equal weights~$\forall \pi_b = \frac{1}{B}$. Then, the next state of the system during iterative trajectory rollouts can be defined as:
\begin{equation}
    \vx_{t+1} = \vx_{t} + {\mathbb{E}}\left[\vE_{b}(\vx_t, \vu_t; \theta_{b})\right] \Delta{t} \,, \quad b \sim \{1, \dots , B\} \,.
    \label{eq:dynamics}
\end{equation}

\subsection{Model Predictive Active Exploration \label{sec:method_active}}
To boost the model learning process, we introduce an active exploration strategy in the \textit{training phase}. We encourage the robot to seek automatically to visit currently unforeseen state-action pairs for training data. To this end, ensemble model disagreement~$D$ can be used as exploration bonuses~\cite{shyam_model-based_2019}. One possible solution is to measure the empirical variance over the predictive samples~$\tilde{\dot{\vx}}_{t+1}$ from the posterior distribution of each ensemble member:
\begin{equation}
    \mathfrak{U}(\vx_t, \vu_t) = \operatorname{Var}\left(\left\{ \tilde{\dot{\vx}}_{t+1} \sim \vE_{b}(\vx_t, \vu_t) \right\}_{b=1}^{B}  \right) ,
    \label{eq:uncertainty_sampling}
\end{equation}
%\Db(\vx_t, \vu_t) \triangleq 
which is referred as \textit{uncertainty sampling}. However, this method cannot distinguish epistemic uncertainty from aleatoric uncertainty. Importantly, the robot should only be inquisitive about the epistemic uncertainty and not the aleatoric uncertainty; otherwise, it will be obscure if this exploration bonus resulted from a lack of knowledge or from inherent and irreducible system stochasticity, such as model ambiguity or sensor noise~\cite{schultheis_receding_2019}.

Kullback–Leibler~(KL) divergence is the most well-known metric for measuring the disagreement between distributions to obtain epistemic uncertainty. However, KL divergence is asymmetric and can only be used to determine the distance between two distributions. Therefore, we employ Jensen-Rényi Divergence~(JRD)
\begin{equation}
    \operatorname{JRD}\left( \vE_{1:B}; \alpha \right) \triangleq  H_{\alpha}\left(\sum_{b=1}^{B} \pi_{b} \vE_{b} \right)- \sum_{b=1}^{B} \pi_{b} H_{\alpha}\left(\vE_{b}\right) ,
\end{equation}
where the Rényi entropy~\cite{renyi_measures_1961} of a random variable~$X$ and its corresponding density function~$p(x)$ is defined as:
\begin{equation}
    H_{\alpha}(X \mid x \in X)=\frac{1}{1-\alpha} \log \int p(x)^{\alpha} \mathrm{d} x \,.
\end{equation}
While an approximation of the JRD can be obtained via Monte-Carlo sampling, it is computationally expensive, thereby precluding real-time implementation. To circumvent this problem, Wang et al.~\cite{wang_closed-form_2009} introduce a closed-form JRD with quadratic entropy~($\alpha = 2$) for analytical measurement of the divergence of a GMM:
\begin{equation}
\begin{aligned}
    D(\vx_t, \vu_t; \vE_{1:B}) = - \log \Biggl[ & \frac{1}{B^{2}}  \sum_{i, j}^{B} \mathfrak{D}\left(\vE_{i}, \vE_{j}\right)\Biggr] + \\
    & \frac{1}{B} \sum_{i}^{B} \log \left[ \mathfrak{D}\left(\vE_{i}, \vE_{i}\right) \right] ,
    \label{eq:jrd}
\end{aligned}
\end{equation}
where
\begin{equation}
    \mathfrak{D}\left(\vE_{i}, \vE_{j}\right)=\frac{1}{|\bm{\Phi}|^{\frac{1}{2}}} \exp \left(-\frac{1}{2} \bm{\Delta}^{\top} \bm{\Phi}^{-1} \bm{\Delta}\right),
\end{equation}
$\bm{\Phi}=\VSigma_{\theta_{i}}+\VSigma_{\theta_{j}}$ and $\bm{\Delta}=\vmu_{\theta_{i}}-\vmu_{\theta_{i}}$.

Using this metric, Shyam et al.~\cite{shyam_model-based_2019} presented a novel active exploration strategy by measuring the ensemble model disagreement along the future predictions. The proposed method trained a model-free policy to harness this time-varying and intricate exploration bonus. However, due to the sample inefficiency and the inability to adjust the trained policy to achieve modified task-specific objectives, this method has difficulty being used to robotic applications. To make these limitations tractable, we here employ Model Predictive Path Integral~(MPPI) control~\cite{williams_information_2017} as a policy.

MPPI is the state-of-the-art sampling-based MPC that enables real-time implementation with the aid of Graphic Processing Units~(GPUs) due to its parallelizable structure~\cite{williams_information-theoretic_2018}. Denoting by $U = \{\vu_0, \dots, \vu_{T-1}\}$ with a fixed time horizon~$T$, the MPPI algorithm seeks an optimal control sequence~$U^{*}$ such that:
\begin{equation}
    U^{*} = \underset{U}{\operatorname{argmin}} \, \mathbb{E} \left[\phi(\vx_{T})+\sum_{t=0}^{T-1} \mathcal{L}\left(\vx_t, \vu_t\right) \right] ,
\end{equation}
where $\phi\left( \cdot \right)$ is a state-dependent terminal cost. In general, the running cost with the quadratic control cost with control weight matrix $\vR$ takes the form:
\begin{equation}
    \mathcal{L}\left(\vx_t, \vu_t\right) = q \left( \vx_{t} \right) + \frac{1}{2} \vu_{t}^{\top} \vR \vu_{t} \,.
    \label{eq:running_cost}
\end{equation}
Note that the state-dependent running cost~$q\left( \cdot \right)$ can be arbitrary functions, allowed by the sampling-based optimization scheme that is capable of solving non-convex problems. Consequently, we can include the information gain obtained from the model disagreement~(\ref{eq:jrd}) into the controller's objective~(\ref{eq:running_cost}) in order to jointly achieve task-dependent cost optimization and active exploration:
\begin{equation}
    \mathcal{L}_{\text{active}}\left(\vx_t, \vu_t\right) = q \left( \vx_{t} \right) + \frac{1}{2} \vu_{t}^{\top} \vR \vu_{t} - w_{D} \, D(\vx_t, \vu_t) \,,
    \label{eq:jrd_cost}
\end{equation}
where $w_{D} > 0$ is a weighting constant. As a result, the robot is encouraged to choose behaviors that lead to exploring uncertain state-action spaces of the dynamic model, while still attempting to minimize the task-specific error. 

The above characteristic facilitates the robotics applications of this method in three ways. First, the controller actively seeks states and actions having high epistemic uncertainty, i.e., for which there are little training data, in the vicinity of the set task objective. Thus, the trained model is completely tailored for instantaneous deployment. Second, it is conceivable to determine when the \textit{exploration phase} may be terminated; specifically, it is when the robot optimizes mostly for the task cost. As training data accumulate, the ensemble models will gradually converge toward increasingly similar predictions, resulting in a progressive decrease in model disagreement. Finally, the robot will be able to adapt to the changes in task costs and to solve the modified tasks without re-training.

\subsection{Uncertainty-Aware Deployment}

After sufficient training time, the robot may be deployed to perform some designated tasks. In the \textit{deployment phase}, we generally regard that there are no human supervisors and no further dynamics update. Recent research has shown that parallel ensemble neural networks are capable of learning dynamics online~\cite{jiahao_online_2022}. However it is orthogonal to the goal of this paper, and our method also can benefit from this consideration. For successful operations, the robot must avoid unpredictable situations, which can be accomplished by incorporating uncertainty into safety violations, regardless of the source of uncertainty. For instance, a data point may have no disagreement across ensembles due to sufficient training data~(low epistemic uncertainty), but we must still avoid this point if the variance of the predicted posterior distribution is high~(high aleatoric uncertainty).

In this phase, we use a variant of the MPPI algorithm called Smooth MPPI~(SMPPI)~\cite{kim_smooth_2022}. SMPPI shares the same information-theoretic roots as MPPI, and it also benefits from the structure of parallel trajectory evaluation. However, SMPPI lifts the control variables as derivative actions, so that the noisy sampling is performed on a higher order domain~$\dot{U} = \{\dot{\vu}_0, \dots, \dot{\vu}_{T-1}\}$. Let us denote the resulting optimal control trajectory after the optimization process as $\dot{U}^{*}_{i+1}$ at MPC iteration step~$i$. Then, the optimal action trajectory~$U^{*}_{i+1}$, which will be used as the actual commands sent to the robot, is obtained by the integral with respect to the MPC iteration horizon: $U^{*}_{i+1} = U^{*}_{i} + \dot{U}^{*}_{i+1} \Delta i$, where $\Delta i$ represents the control updating period of MPC and is commonly equal to $\Delta t$. This new action trajectory update law allows the controller to rapidly respond to changing environments while using significantly lower sampling variance, thus alleviating the chattering in resulting commands. Furthermore, we can apply an extra action smoothing cost~$\Omega(U)$ along trajectory rollouts:
\begin{equation} \label{eq:actionseq}
    \Omega(U) = \sum_{t=1}^{T-1} (\vu_t - \vu_{t-1})^{\top} {\bm{\omega}} \, (\vu_t - \vu_{t-1}) ,
\end{equation}
where $\bm{\omega} \in \mathbb{R}^{m \times m}$ is the weighting parameter. Such action cost was not eligible in the MPPI baseline with non-affine dynamics because it violates the information theoretic derivation~\cite{kim_smooth_2022}. As a result of these smoothing effects, SMPPI is not suitable for rapid exploration in underactuated continuous control systems but is beneficial for deployment.

To achieve risk-averse planning, taking into account both aleatoric and epistemic uncertainty, we employ the results of \textit{uncertainty sampling}~(\ref{eq:uncertainty_sampling}) as uncertainty measurements. A simple addition of the quantified uncertainty to the cost function would make the controller uncertainty-aware, as the following form:
\begin{equation}
    \mathcal{L}_{\text{naive}}\left(\vx_t, \dot{\vu}_{t}, \vu_t\right) = q \left( \vx_{t} \right) + \frac{1}{2} \dot{\vu}_{t}^{\top} \vR \dot{\vu}_{t} \,+ w_{1}  \mathfrak{U}(\vx_t, \vu_t) ,
    \label{eq:naive_penalty}
\end{equation}
where $w_{1} > 0 $ is a weighting constant. It is important to point out that the measured uncertainty through sampling is not differentiable with respect to state and action. Sampling-based MPC is capable of optimizing any arbitrarily crafted cost functions, making such a form of uncertainty tractable. Similarly, we take use of this trait by imposing an impulse-like penalty where uncertainty exceeds a threshold~$\xi$:
\begin{equation}
\begin{aligned}
    \mathcal{L}_{\text{hybrid}}\left(\vx_t, \dot{\vu}_{t}, \vu_t\right) = q \left( \vx_{t} \right) & + \frac{1}{2} \dot{\vu}_{t}^{\top} \vR \dot{\vu}_{t} \,+ w_{1} \mathfrak{U}(\vx_t, \vu_t) + \\
      & w_{2} \, I \Bigl( \Bigl\{ \Bigl|  \mathfrak{U}(\vx_t, \vu_t) \Bigr| > \xi \Bigr\} \Bigr) ,
\end{aligned}
\label{eq:hybrid_penalty}
\end{equation}
where $w_{2} \gg w_{1} $ is a weighting constant and $I$ is an indicator function. It is an intuitive consideration because we never want to allow the robot to take actions that are completely unpredictable.

Following is the final objective function of SMPPI:
\begin{equation}
\begin{aligned}
    \dot{U}^{*} = \underset{\dot{U}}{\operatorname{argmin}} \, \mathbb{E} \Biggl[\phi(\vx_{T}) + & \Omega( U + \dot{U}\Delta i) \\
    & + \sum_{t=0}^{T-1}\mathcal{L}_{\text{hybrid}}\left(\vx_t, \dot{\vu}_{t}, \vu_t\right) \Biggl] .
\end{aligned}
\end{equation}

\subsection{Implementing Neural Network Vehicle Dynamics \label{sec:method_implementing}}

Given that we have made no assumption on the specific robotic platform in our framework, we argue that it is applicable to any kind of robotic system. In this work, we focus on autonomous vehicle and wheeled robot applications to validate our idea. We choose the robot's state and action input based on the dynamic bicycle model since it provides a good trade-off between model accuracy and simplicity for real-time implementation \cite{kabzan_learning-based_2019}. The dynamic state variable~$\vx = (v_x, v_y, r)^{\top}$ consists of longitudinal velocity, lateral velocity, and yaw rate. These states can be easily measured by using GPS and IMU sensors. The action input~$\vu = (\delta, v_{des})^{\top}$ consists of the steering angle and desired speed. The desired speed is sent to the low-level Proportional-Integral (PI) controller, which determines the throttle and brake.

Since these state and action variables are simplified representations of the full robot dynamics, there exists irreducible model ambiguity. For example, the roll and pitch motions are not considered in the bicycle model. Automatic gear shifting encompasses non-smooth and time-varying dynamics characteristics, hence increasing the lower bound of model bias. We alleviate this problem by providing the history of state-action pairs to the neural network input so that it can extract contextual information~\cite{spielberg_neural_2019, kim_toast_2022}. The history length~$H$ must be carefully determined, because as $H$ increases, the state-action space that has to be discovered becomes exponentially larger, making the exploration problem more difficult. We select $H=4$ in this work through an ablation study in the history length (see Section~\ref{sec:appendix}). Although LSTM~\cite{hochreiter_long_1997} and GRU~\cite{cho_properties_2014} show slightly better prediction performances than Multi-Layer Perceptron~(MLP) (see Section~\ref{sec:appendix}), we choose MLP since its operations are fully parallelizable.

We take the parallel ensemble MLP implementation by \textit{philipjball}~\cite{baddbmm} and adapt it for our task. Let us denote the weight and the bias of a layer of an ensemble model~$\vE_{b}$ as $\vW_b$ and $\vb_b$, respectively. Then, the parameters of all ensemble models can be represented in the form of batches of matrices: $\vW = [\vW_1^{\top}, \dots, \vW_B^{\top}]^{\top}$, $\vb = [\vb_1, \dots, \vb_B]^{\top}$. First, we initialize the parameters of each ensemble model according to the size of the layer input~$n_{\text{in}}$:
\begin{equation}
    \vW_b, \, \vb_b \sim \mathcal{U} \left(-\sqrt{\frac{1}{n_{\text{in}}}}, +\sqrt{\frac{1}{n_{\text{in}}}}\right) , \forall b \in \{1, \dots, B\} ,
\end{equation}
where $\mathcal{U}$ is uniform distribution. While training the model with collected data, the forward propagation of each ensemble model is computed through
\begin{equation}
    \vo_b = \vW_b^{\top} \vz + \vb_b , \quad \forall b \in \{1, \dots, B\}, 
\end{equation}
where $\vz$ and $\vo$ are the input and the output of the layer, respectively. The gradients of the Gaussian NLL loss function~(\ref{eq:train_loss}) can be obtained through automatic differentiation packages such as Pytorch \texttt{autograd}~\cite{paszke_automatic_2017}. This process is identical to the common way we use MLP. While using the model as the dynamics of the robot, on the other hand, the forward propagation of the entire ensemble model is defined as batch matrix-matrix multiplication and addition:
\begin{equation}
    [\vo_1, \dots, \vo_B]^{\top} = \vW^{\top} \odot [\vz, \dots, \vz]^{\top} \oplus \vb ,
    \label{eq:baddbmm}
\end{equation}
where $\odot$ and $\oplus$ are element-wise operations. This parallelized forward propagation~(\ref{eq:baddbmm}) can be implemented using Pytorch \texttt{baddbmm} function. Therefore, with the aid of sufficient GPU resources, we can simultaneously predict the next state and obtain uncertainty using PENN dynamics in a fixed-time duration regardless of the number of ensembles. This strategy maximizes the benefits of the parallel sample evaluation property of the MPPI framework.

\section{Experiments \label{sec:exp}}

Our experimental evaluations address the following key questions: \textbf{(Q1)} Can our method explore the state-action space with sufficient sample efficiency~(see Section~\ref{sec:exp_active})? \textbf{(Q2)} Can the PENN taught with active exploration be used for long horizon planning, even if the task objective has been modified after training~(see Section~\ref{sec:exp_direct})? \textbf{(Q3)} Can our method be turned directly into an uncertainty-aware controller with minimal modification~(see Section~\ref{sec:exp_safe})? \textbf{(Q4)} Is our method scalable to other robot platforms~(see Section~\ref{sec:exp_wheeled})?

We evaluate \textbf{Q1}-\textbf{Q3} in a high-fidelity vehicle dynamics simulator~-~IPG CarMaker. We build the modeling and simulation system by integrating the algorithms using ROS~2~\cite{macenski_robot_2022}. A Land Rover Defender 110 is used as the control vehicle. The low-level PI controller runs at 100~Hz, and the control parameters~$K_{\text{P}}, K_{\text{I}}$ are set to \{0.35, 1.0\} for throttle and \{0.18, 0.3\} for brake. To simulate sensor noise, we add 10\% of i.i.d. Gaussian noise to each dynamic state observation in $\vx$. MPPI is used in the active exploration experiments~\textbf{(Q1)}, and SMPPI is used in the deployment experiments~\textbf{(Q2-Q3)}. Table~\ref{table:mppi} lists the parameters of two controllers, while the remaining parameters not specified in this paper are taken from the SMPPI implementation~\cite{kim_smooth_2022}.

\begin{table}[htb!]
\caption{Control parameters of MPPI and SMPPI. Note that the action smoothing cost with $\bm{\omega}$ only applies to SMPPI.}
\renewcommand{\arraystretch}{1.2}
\begin{center}
\begin{tabular}{c|c|c}
\qquad Parameters \qquad\qquad &  MPPI~\cite{williams_information-theoretic_2018} \qquad &  SMPPI~\cite{kim_smooth_2022} \qquad\\ \hline
$\Delta t$ & 0.1 s & 0.1 s\\
$\Delta i$ & 0.1 s & 0.1 s\\
$K$ &  5,000 & 5,000\\ 
$T$ & 10 & 35\\ 
Sampling Variance & ${\bf{\text{Diag}}}(4.0, 3.0)$ & ${\bf{\text{Diag}}}(1.6, 0.4)$\\
$\bm{\omega}$ & N/A & ${\bf{\text{Diag}}}(0.4, 0.1)$ 
\end{tabular}
\end{center}
\label{table:mppi}
\end{table}

We use a four-layer MLP  with the hidden layer sizes \{40, 80, 120, 40\} throughout all experiments. Rectified Linear Unit~(ReLU)~\cite{nair_rectified_2010} is used as activation functions. Similar to \cite{chua_deep_2018}, we use five ensembles~($B = 5$) for real-time implementation.
\subsection{Model Predictive Active Exploration \label{sec:exp_active}}

\textbf{Experimental Setup.}
In this experiment, we evaluate the effectiveness of the active exploration strategy~\textbf{(Q1)}. We let the robot seek for training data on a large flat ground with 0.7 friction coefficient. The task objective is to maintain the 50~km/h target speed~($v_{\text{target}}$) while staying inside the space boundary. Accordingly, the task-dependent running cost~$q(\vx_t)$ is defined as follows:
\begin{equation}
\begin{aligned}
    q(\vx_t) & = \alpha_1{\text{Track}(\vx_t)} + \alpha_2{\text{Speed}(\vx_t)}  \,,\\
    \text{Track}(\vx_t) & = (0.9)^t\,{10000 \, \textbf{M}(p_x,p_y)} \,, \\
    \text{Speed}(\vx_t) & = \left(\sqrt{{v_x}^2 + {v_y}^2} -v_{\text{target}}\right)^2 .
\end{aligned}
\end{equation}
$\alpha_{(\cdot)}$ are empirically tuned weighting constants. The track cost imposes a hard penalty to prevent collisions. The given map~$\textbf{M}$ indicates whether the robot position~$(p_x,p_y)$ is at the outside of the space boundary.

We assume that there are human monitors who supervise the \textit{exploration phase}. They disengage the robot's autonomy before the robot reaches the space boundary or before the robot enters dangerous situations. We empirically assume this period as 30~s. While the human supervisors setting up the robot for a new trial, the PENN is trained with the collected data offline. We refer to this process as one \textit{iteration} for exploration. 

\textbf{Methods Evaluated.}
We evaluate the following methods including ours:
\subsubsection{Ours} a model predictive active exploration strategy, which optimizes the objective function of (\ref{eq:jrd_cost}). The epistemic uncertainty measured by JRD~(\ref{eq:jrd}) serves as the information gain. We set the information gain weighting constant~$w_{D}$ as 100.
\subsubsection{Uncertainty Sampling~(US)} an uncertainty sampling counter-part of our method, which uses the uncertainty quantified through sampling over each ensemble posterior distribution as the information gain. The substitution of JRD~(\ref{eq:jrd}) with uncertainty sampling~(\ref{eq:uncertainty_sampling}) in (\ref{eq:jrd_cost}) is used as its running cost. We use the same weighting~$w_{D} = 100$ as our method since the quantified information gain in both methods is of similar magnitude in our implementation.
\subsubsection{Jensen-Rényi Divergence Reactive Exploration~(JDRX)} a reactive counter-part of our method~\cite{shyam_model-based_2019}. We simply implement this based on our method by setting the planning horizon to $T=1$, which explores greedily without planning based on the experience collected so far.
\subsubsection{Random Noise~(RN)} a random exploration strategy, which injects noise in the action input. Its objective function is (\ref{eq:running_cost}) and there is no information gain. We sample random noise from the Gaussian distribution with the same variances as the sampling distribution used by MPPI.

All the compared methods are evaluated with the same control parameters and the same model training setting with a fixed random seed. While training the model offline, we randomly split the collected data~$\calD$ into the training and test sets with a 7:3 ratio. The PENN is trained using the Adam optimizer with a learning rate of 0.001. We save the model that shows the best test sets performance during 10~epochs, and use this model as the robot dynamics~$\vF$ in the next iteration.

\begin{wrapfigure}{R}{0.5\linewidth}
\centering
\includegraphics[width=0.99\linewidth]{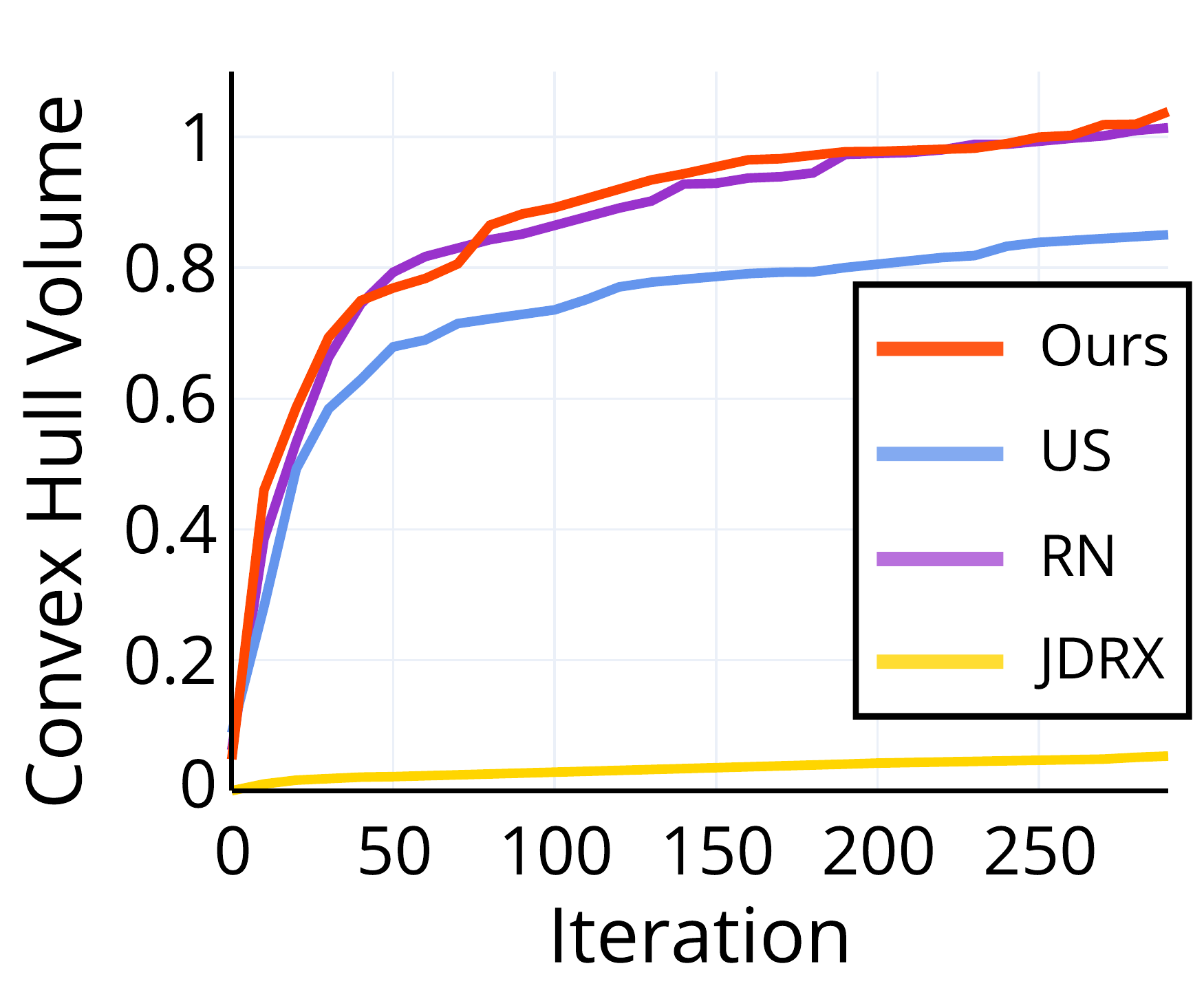}
\caption{The volumes of the convex hulls along exploration iterations.}
\label{fig:volume}
\end{wrapfigure}

\begin{figure*}[t]
\centering
\includegraphics[width=0.99\textwidth]{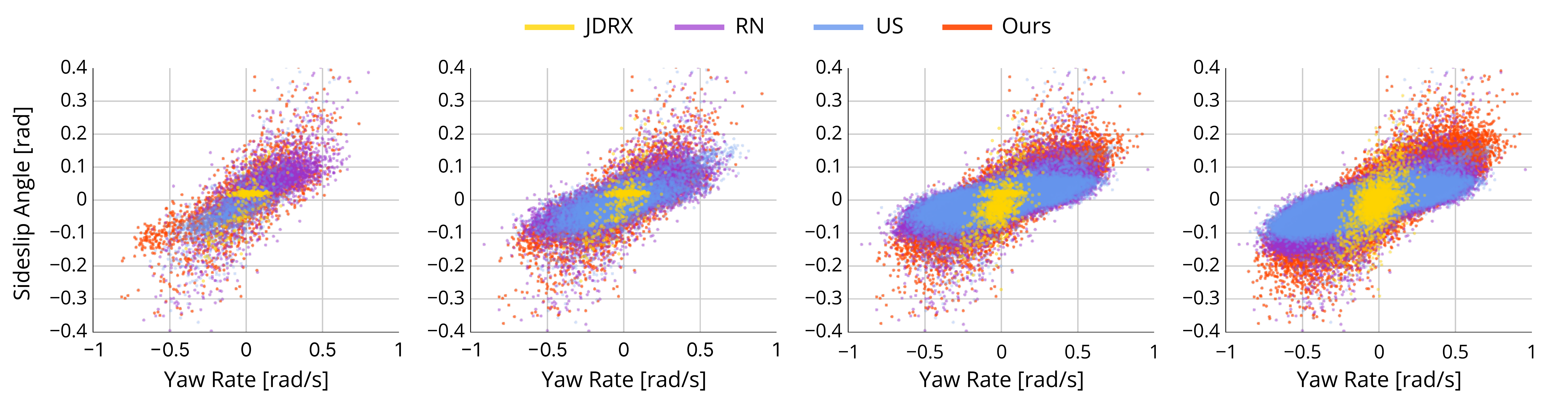}
\caption{The scatter plots of collected data during active exploration. The data are from 10, 30, 100, and 300 iterations, respectively. During all iterations, ours using JRD information gain covers the largest state spaces of sideslip angle and yaw rate compared to other methods.}
\label{fig:coverage}
\vspace*{-0.05in}
\end{figure*}

\textbf{Experimental Results.}
We first evaluate how much of the state-action space each method has covered. We compute the convex hull~\cite{barber_quickhull_1996} with the collected state and action data~$\{\vx_t, \vu_t\}_{t=0}^{N-1}$ at each iteration, and analyze the volume enlargement of the convex hulls~\cite{liu_towards_2021}~(see Fig.~\ref{fig:volume}). The results indicate that JDRX cannot explore the state-action space efficiently. Also, the volumes of ours and RN show comparable magnitudes while consistently exceeding those of US.

We further analyze the collected data points in detail. Similar to as Spielberg et al.~\cite{spielberg_neural_2021}, we focus on the state envelope of sideslip~$\beta$ and yaw rate~$r$, since precise motion predictions with large sideslip angles and rotational speed are crucial to the success of high-speed driving on sharp curves. The results are shown in Fig~\ref{fig:coverage}. JDRX is not able to collect driving data with high yaw rates. This is because it reacts immediately to the trivial information gain near the current state and is unable to plan for higher information gain in the future. On the other hand, ours always covers the largest state space among the compared methods. This is because ours can plan over a longer horizon to gather drifting data by ignoring the aleatoric uncertainty exudes from plain driving. Our method has collected a substantial amount of data with $\beta$ around 0.2~rad~(approximately 11.5~$^\circ$), and even data with $\beta$ around 0.3~rad~(approximately 17.2~$^\circ$). Lastly, we stress that the data collected by RN is not helpful for achieving the driving tasks, despite it showing a similar enlargement in the convex hull to our method. We will further discuss this in Section~\ref{sec:exp_direct}.

\begin{figure*}[t]
\centering
\subfloat[]{\includegraphics[width=0.45\textwidth]{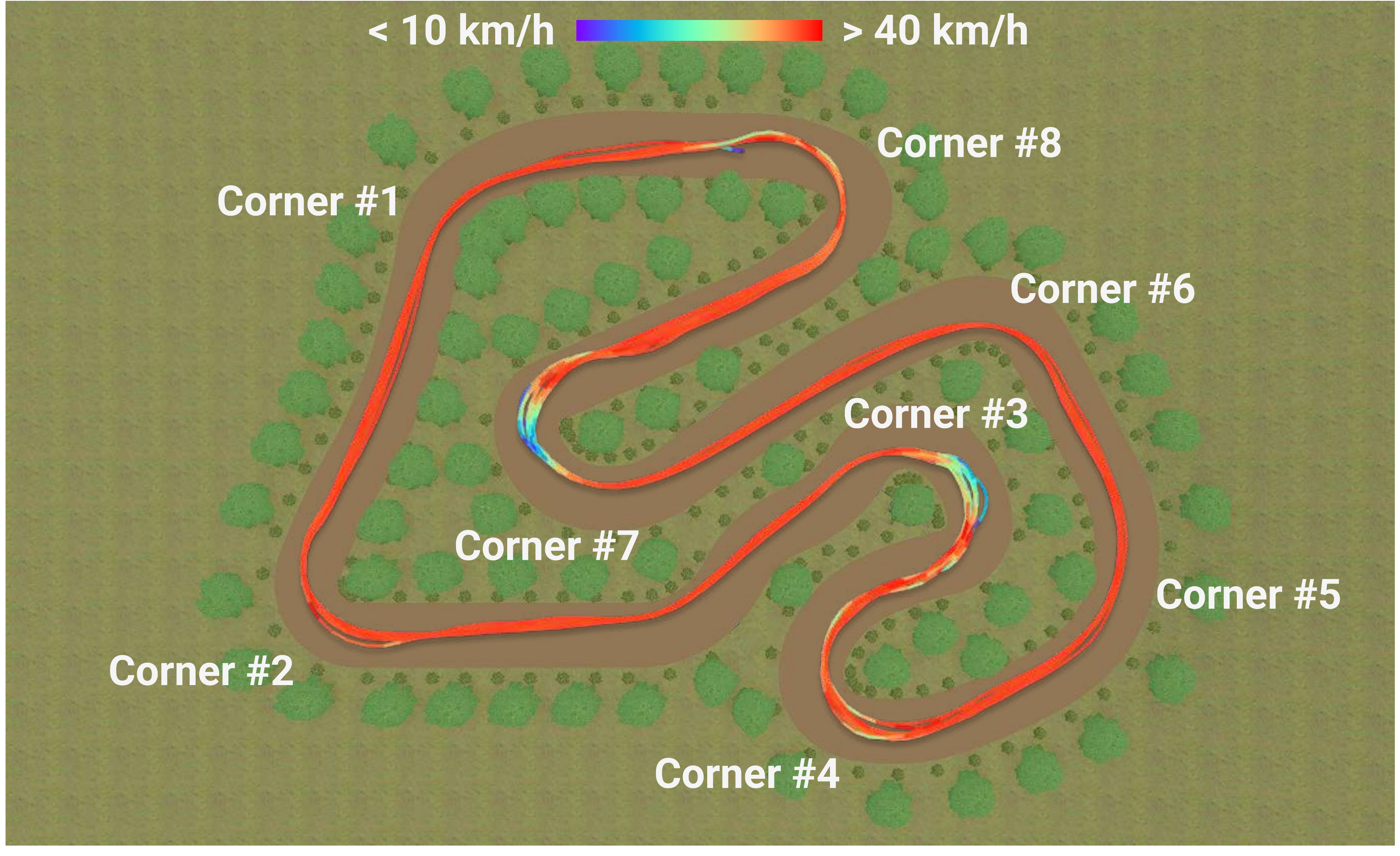}\label{fig:cm_map}}
\subfloat[]{\includegraphics[width=0.45\textwidth]{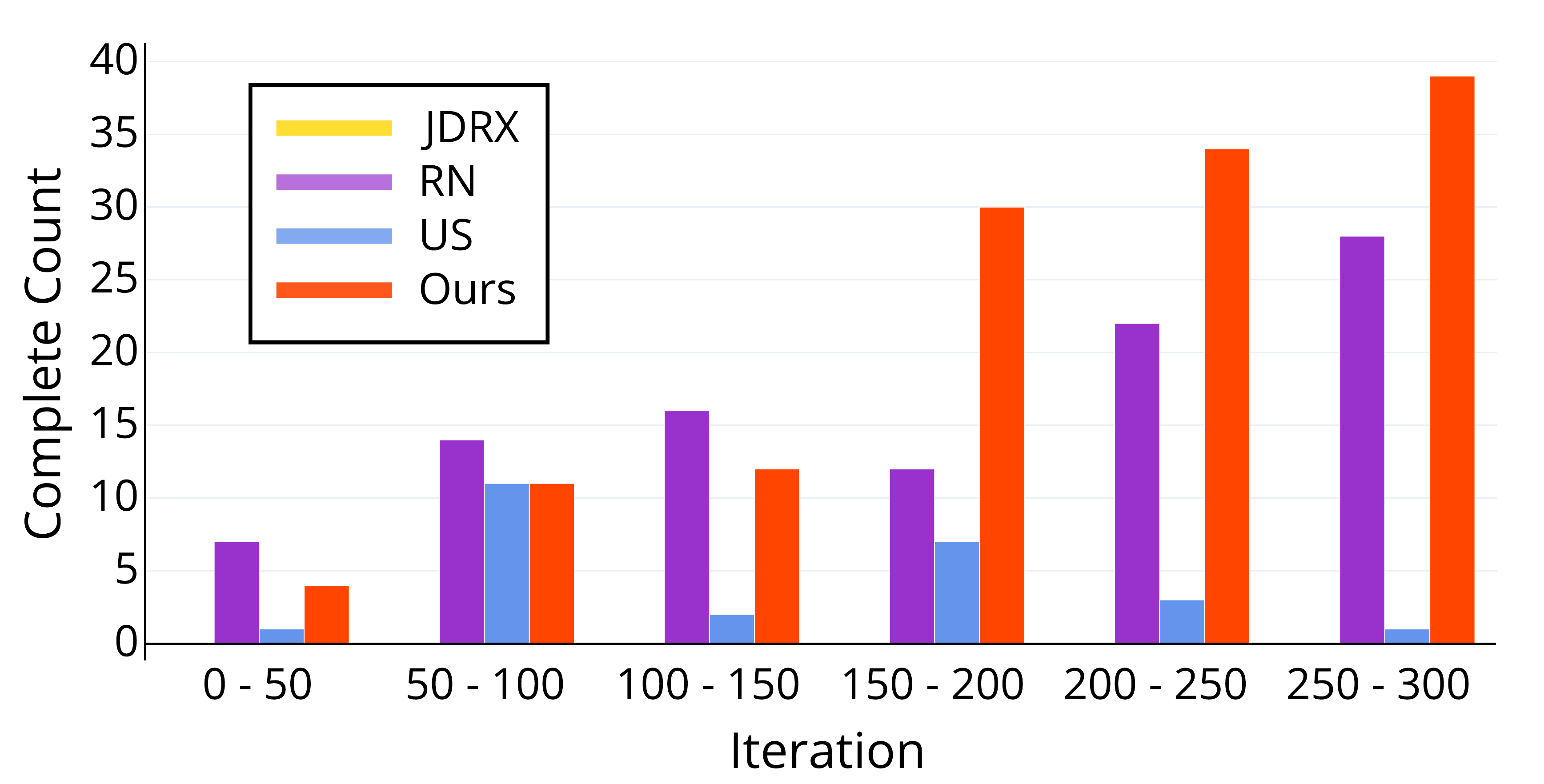}\label{fig:complete}} \hfill
\subfloat[]{\includegraphics[width=0.45\textwidth]{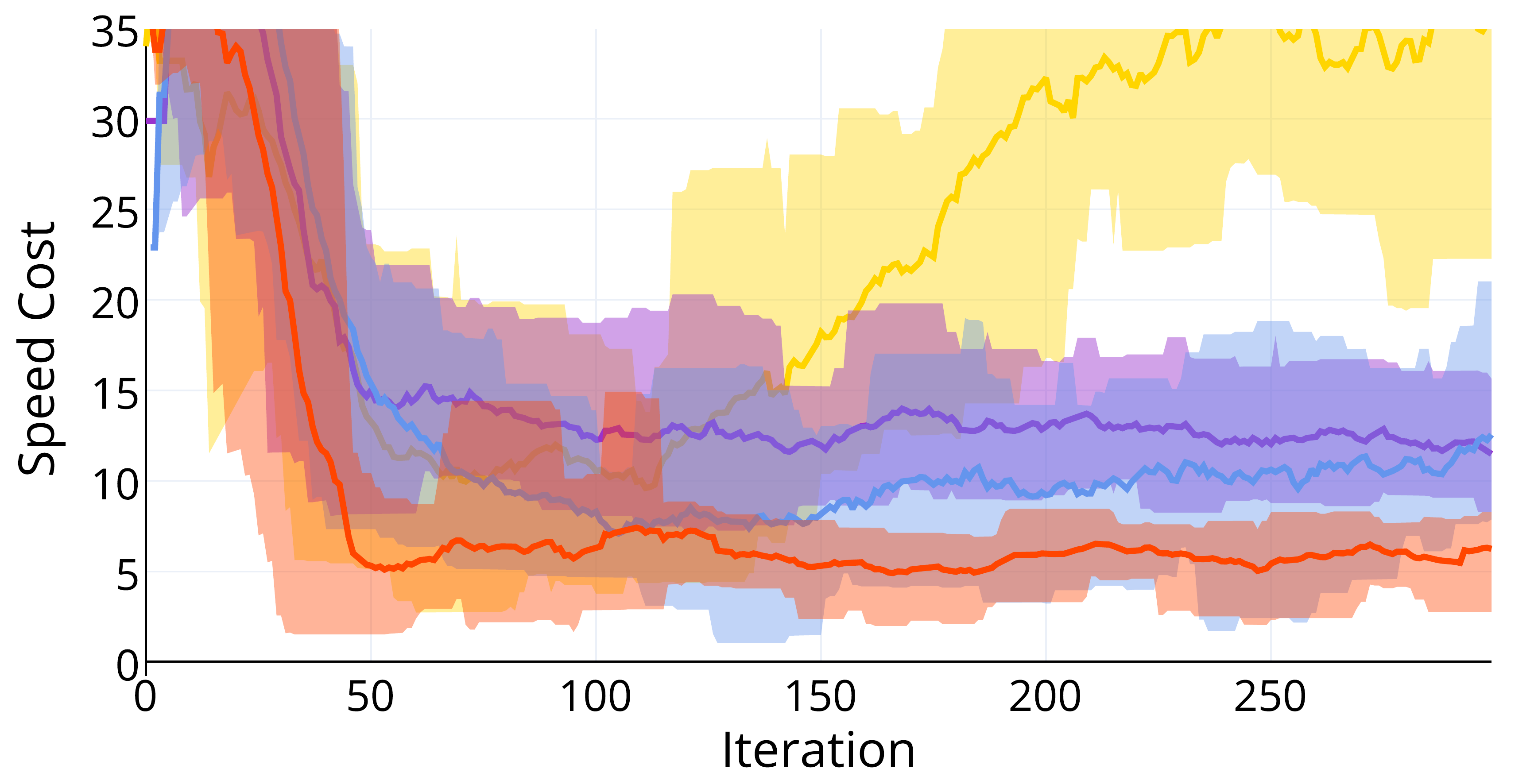}\label{fig:speed}}
\subfloat[]{\includegraphics[width=0.45\textwidth]{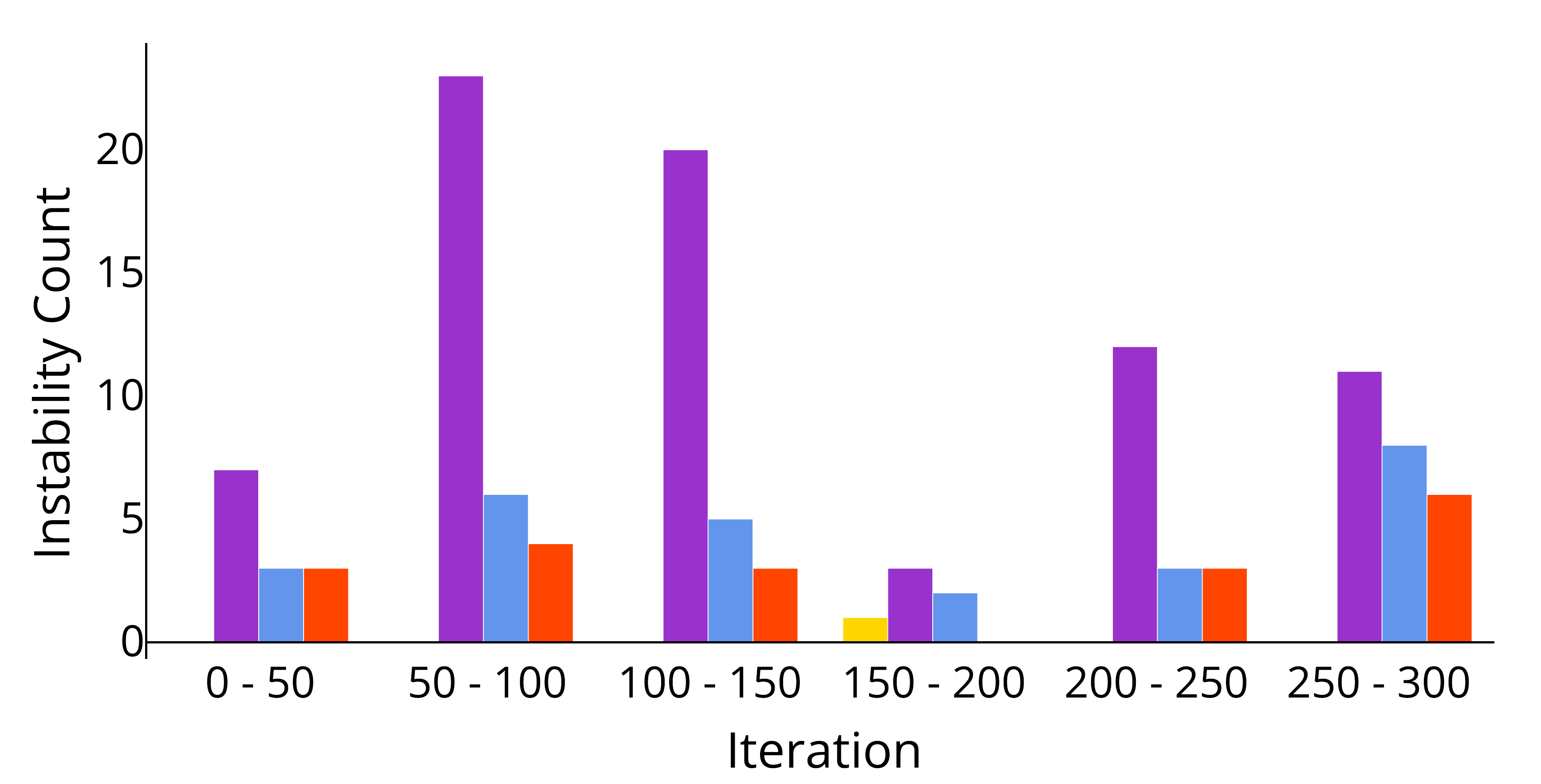}\label{fig:instability}}
\caption{
    (a) The race track designed in the IPG CarMaker simulator for direct deployment experiments. We visualize the vehicle trajectory taken by our method at 300~iterations for driving 10~laps in a counter-clockwise direction. 
    (b) The number of times that each method completes the whole lap during every 50~iterations. 
    (c) The average speed cost for each saved model along exploration iterations. The shaded areas denote a 95\% confidence interval.
    (d) The number of trials of each method that violates the stabilizing constraints at least once, i.e., having larger sideslip angles than 0.3~rad, during every 50~iterations.
}
\label{fig:direct_inference}
\end{figure*}
\subsection{Direct Deployment \label{sec:exp_direct}}

\textbf{Experimental Setup.}
We take the resulted PENNs of active exploration as the robot dynamics and directly evaluate their performances in long horizon planning~\textbf{(Q2)}. We load the best models at each iteration and monitor the control performance improvements of the PENNs of the compared methods. The performances are evaluated in a race track with three moderate curves and five sharp curves~(see Fig.~\ref{fig:cm_map}). We set the friction coefficient of the track to 0.7.  

In this experiment, the objective of the robot is to drive through the race track at a speed as close to $v_{\text{target}}$ as possible, while staying inside of the track. We modify the task-dependent cost~$q(\vx_t)$ as follows:
\begin{equation}
    q(\vx_t) = \alpha_1{\text{Track}(\vx_t)} + \alpha_2{\text{Speed}(\vx_t)} + \alpha_3{\text{Stable}(\vx_t)}.
    \label{eq:deploy_cost}
\end{equation}
The track cost and the speed cost are nearly identical to those in the exploration phase. The given map~$\textbf{M}$ is modified accordingly to the new race track. The target speed~$v_{\text{target}}$ is adjusted to 45~km/h. We add the stabilizing cost that imposes a hard penalty on large sideslip angles~$\beta$ to prevent the robot from losing its stability:
\begin{equation}
\begin{aligned}
    \text{Stable}(\vx_t) & = 10000 \, I \left( \left \{\left | \beta \right | > 0.3 \right\} \right) \,, \\
    \beta & = -\text{arctan} \left ( \frac{v_y}{\lVert v_x \rVert} \right ) \,.
\end{aligned}
\end{equation}
We set the planning horizon to $T=35$, which is 3.5~s, to compare the performances in long horizon planning. Note that no information gain is provided to the objective function, so the controllers are uncertainty-neutral.

\textbf{Experimental Results.}
The results are shown in Fig.~\ref{fig:direct_inference}. At the sharpest corners of $\#3$, $\#7$, and $\#8$, in order to drive through them without losing significant speed, the vehicle should gently slow down before entering the corners and drift continuously. If the vehicle cannot perform drifting maneuvers, it will nearly stop in the middle of the corners or even collide with the track boundary. Moreover, if the predictions of the dynamics are not accurate, the vehicle may enter a saturated drifting region and become uncontrollable.

First, we count the number of times that each method completes the whole track without driving outside of the given boundary~(see Fig.~\ref{fig:complete}). JDRX is not able to complete the whole lap. Reactive exploration limits the robot to acquiring only insignificant information gain in the vicinity of the current state. As a result, it fails to collect high-speed driving data for training and to produce accurate predictions in states with high speeds. RN shows improvement as it collects more data through repeated iterations. However, it only relies on coincidental actions that lead to collecting novel data. Our method achieves the best results among the compared methods, and its performance steadily improves as more data is collected.

Interestingly, the performance of US improves in the early stages, but after 100~iterations, its performance continuously decreases. It implies that collecting more data does not necessarily improve the prediction performance. Since US cannot separate the source of uncertainty, it may repeatedly collect the same data showing naive motions. Therefore, the model is trained to make increasingly better predictions on these superfluous data while neglecting the data representing important driving characteristics.

We also analyze the control performance in terms of speed~(see Fig.~\ref{fig:speed}) and stability~(see Fig.~\ref{fig:instability}) during each trial. Our method has effectively regulated the speed cost compared to the other methods. The results also show that RN violates the stabilizing constraints far more frequently than our method. It demonstrates that the collected data from RN do not effectively contribute to achieving the given task, since they omit crucial training data such as high-speed drifting maneuvers. Only our method successfully drives through the race track at high speeds without losing stability.

\begin{figure*}[t]
\centering
\includegraphics[width=0.99\linewidth]{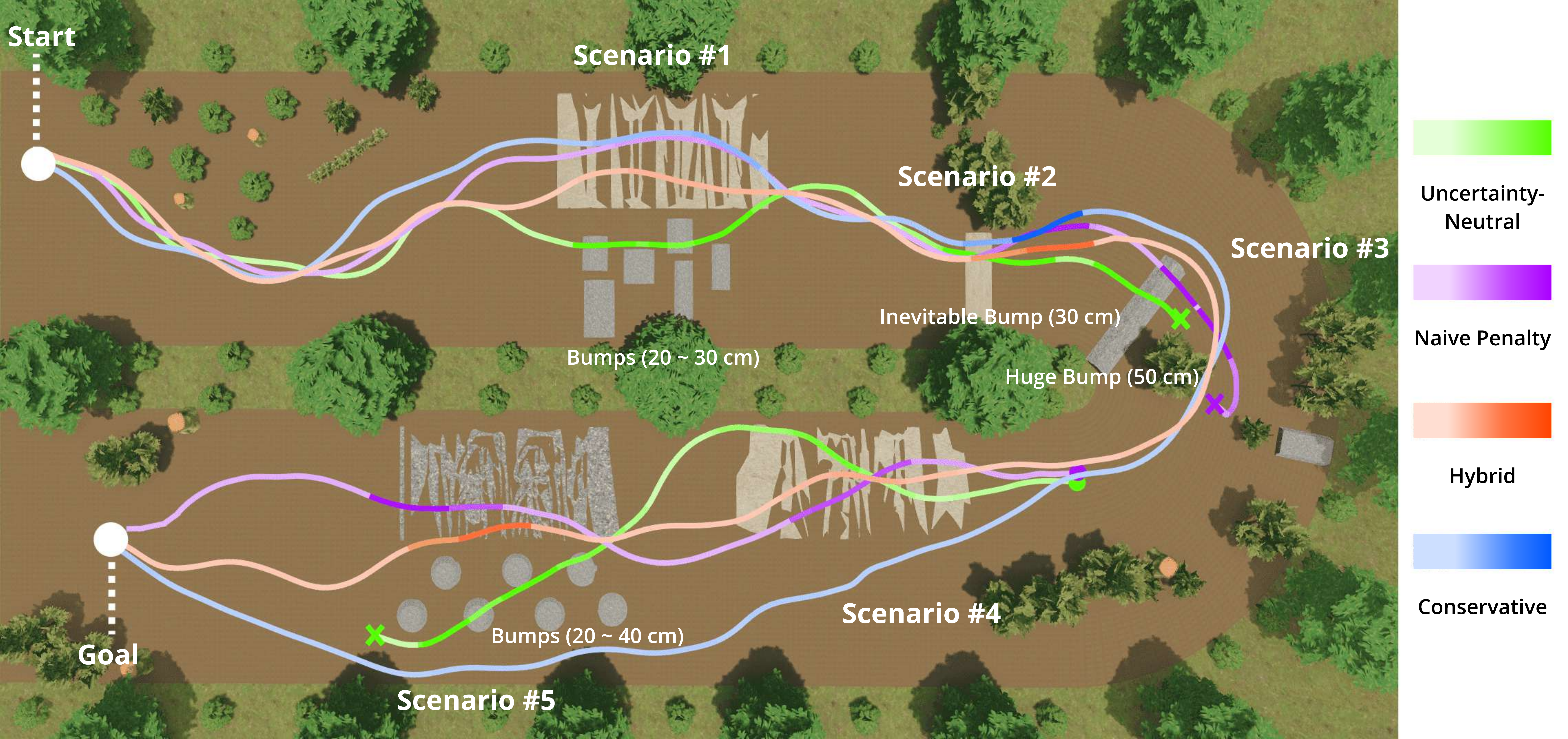}
\caption{Visualization of uncertainty-aware navigation results on the vehicle simulator. We display the rotational impacts exerted on the vehicle onto the trajectories. The maximum value of rotational impact during a three-second window is used for visual clarity.}
\label{fig:safe}
\end{figure*}

\begin{table*}[t!]
\centering
\scriptsize
\caption{The average and maximum motions of the vehicle across all trials. The vertical motions include vertical velocity and vertical acceleration. The angular motions include velocities and accelerations of roll and pitch, respectively. Hybrid and Conservative experience lower terrain impacts than UN and NP by a large margin.}
\resizebox{\textwidth}{!}{
\label{tab:results}
    \begin{tabular}{cccccccccccccc}
        \toprule
        \multirow{2}{*}{\textbf{\#}}  & \multirow{2}{*}{\textbf{Method}} & \multicolumn{2}{c}{\bf{Vertical Vel. [m/s]}} & \multicolumn{2}{c}{\bf{Vertical Acc. [m/s$^\text{2}$]}} & \multicolumn{2}{c}{\bf{Roll Rate [rad/s]}} &
        \multicolumn{2}{c}{\bf{Pitch Rate [rad/s]}} &
        \multicolumn{2}{c}{\bf{Roll Acc. [rad/s$^\text{2}$]}} &
        \multicolumn{2}{c}{\bf{Pitch Acc. [rad/s$^\text{2}$]}} \\
        \cmidrule(l{4pt}r{4pt}){3-4} \cmidrule(l{4pt}r{4pt}){5-6} \cmidrule(l{4pt}r{4pt}){7-8} \cmidrule(l{4pt}r{4pt}){9-10} \cmidrule(l{4pt}r{4pt}){11-12} \cmidrule(l{4pt}r{4pt}){13-14}& & \textbf{Mean} & \textbf{Max} & \textbf{Mean} & \textbf{Max} & \textbf{Mean} & \textbf{Max} & \textbf{Mean} & \textbf{Max} & \textbf{Mean} & \textbf{Max} & \textbf{Mean} & \textbf{Max}\\
        \midrule
        1 & Uncertainty-Neutral (UN) & 2.07 & 28.23 & 0.13 & 1.09 & 0.28 & 2.46 & 0.43 & \textbf{1.10} & 2.22 & 30.67 & 1.64 & 12.78\\
        2 & Naive Penalty (NP) & 1.97 & 23.30 &  0.05 & 0.38 & 0.33 &  2.06 & 0.44 & 2.22 & 2.73 & 33.97 & 1.80 & 8.02\\
        \midrule
        3 & Hybrid & 1.42 & \textbf{9.37} & \textbf{0.04} & \textbf{0.20} & 0.21 & \textbf{1.56} & \textbf{0.39} & 1.15 & 1.85 & \textbf{18.37} & \textbf{1.28} & \textbf{5.91}\\
        4 & Conservative & \textbf{1.40} & 10.60 & \textbf{0.04} & 0.22 & \textbf{0.14} &  1.60 & 0.42 & 1.11 & \textbf{1.16} & 22.23 & 1.46 & 6.93\\
        \bottomrule
    \end{tabular}
    }
\label{table:safe}
\end{table*}
\subsection{Uncertainty-Aware Deployment \label{sec:exp_safe}}

\textbf{Experimental Setup.}
We assumed that the \textit{exploration phase} was performed in a large open space under human supervision. However, in \textit{deployment phase}, there is no such monitoring to guarantee safety. In this experiment, we validate that our active exploration policy can be transferred to an uncertainty-aware controller with minimal modification.

To highlight the influence of the uncertainty-aware control strategy, we augment the dynamics to be terrain-aware by feeding the elevation map to the neural network. We assume that a 2.5D elevation map with a 0.1~m grid size is provided by one of the existing methods that can generate these maps from raw sensor input in real time~\cite{fankhauser_probabilistic_2018,miki_elevation_2022}. The map encoder network consists of an 8-channel convolutional layer and a 4-channel convolutional layer, both with a kernel size of 3 and a stride of 2~\cite{karnan_vi-ikd_2022}. The output of the encoder is concatenated with the history of state-action pairs and fed into the original MLP model that had been used throughout the previous experiments. We use the data collected by our method, which shows the best performance in direct deployment experiments, to train the augmented dynamics. During training, the elevation map is randomly generated from the Gaussian distribution of $\calN(0, 0.2)$. 

Following our prior work~\cite{seo_scate_2023}, we design a realistic off-road environment with various components including large bumps and randomly patterned rough terrains~(see Fig.~\ref{fig:safe}). We also add non-traversable elements such as trees, bushes, and wooden pillars, and they are included in the given map~$\textbf{M}$ for collision avoidance. 

\textbf{Methods Evaluated.}
We take the same task-dependent cost function~(\ref{eq:deploy_cost}) used in the direct deployment experiments. The target speed $v_{\text{target}}$ is adjusted to 30~km/h. The key difference between these two deployment tasks is that we here include uncertainty in the control objective. We evaluate the following uncertainty-averse strategies:
\subsubsection{Uncertainty-Neutral~(UN)} a uncertainty neutral strategy that does not take uncertainty into account while planning. It uses the same objective function as the one used in direct deployment experiments.
\subsubsection{Naive Penalty~(NP)} a simple uncertainty-aware strategy that penalizes a linear cost on quantified uncertainty. It uses (\ref{eq:naive_penalty}) as the objective function with $w_1 = 1000$.
\subsubsection{Hybrid} an uncertainty-aware strategy that has both soft and hard costs. It uses (\ref{eq:hybrid_penalty}) as the objective function with $w_1 = 1000$, $w_2 = 10000$, and $\xi = 0.15$. Note that the hard cost penalizes states that are estimated to have uncertainty greater than $\xi$ with the same magnitude as a collision with the track boundary.
\subsubsection{Conservative} a conservative strategy that controls the vehicle as stable as possible. It uses (\ref{eq:hybrid_penalty}) as the objective function with $w_1 = 0$, $w_2 = 10000$, and $\xi = 0.02$.

\begin{figure*}[t]
\centering
\subfloat[]{\includegraphics[width=0.34\textwidth]{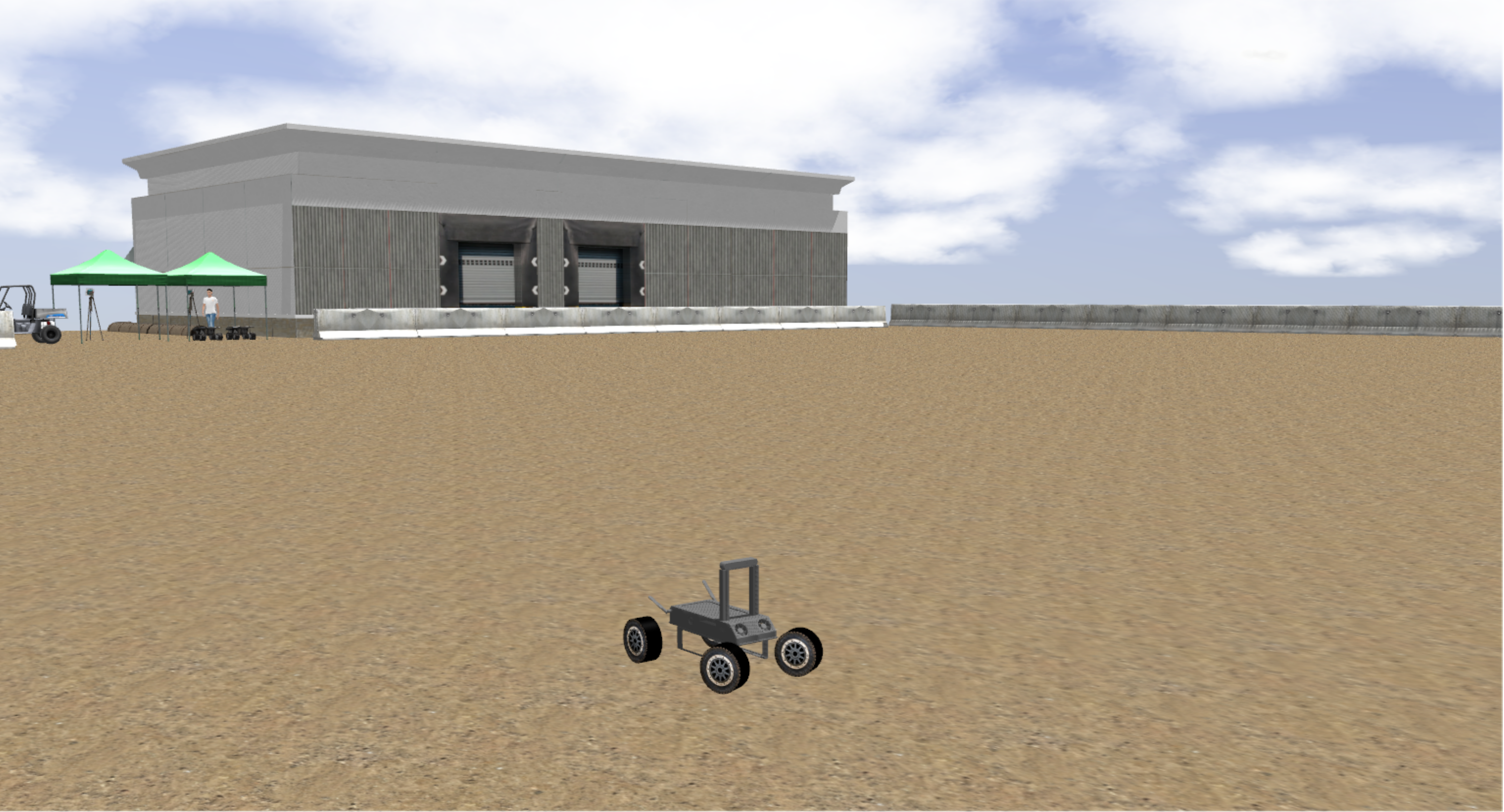}\label{fig:gazebo_explore}}
\subfloat[]{\includegraphics[width=0.34\textwidth]{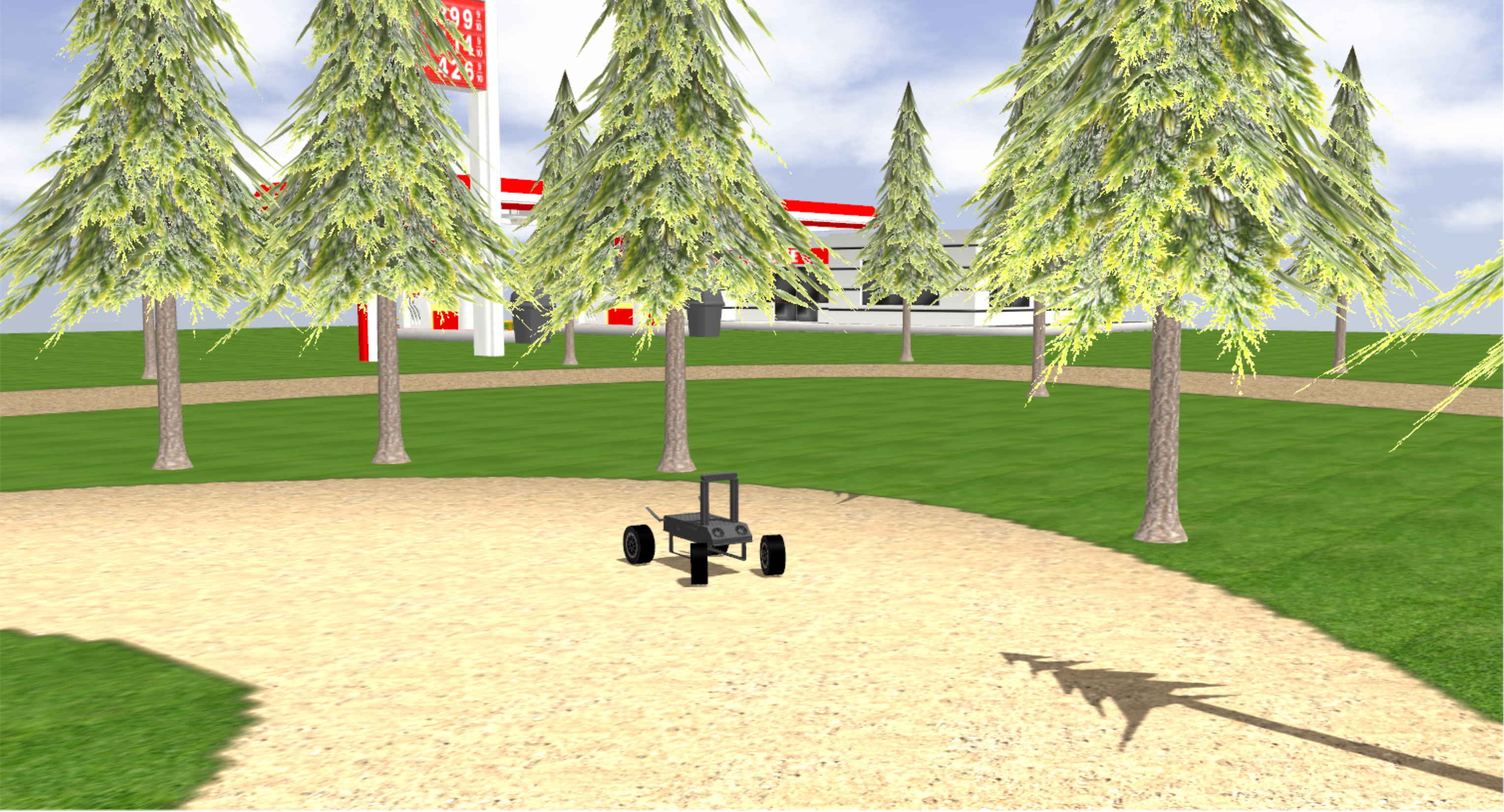}\label{fig:gazebo_deploy}} 
\subfloat[]{\includegraphics[width=0.29\textwidth]{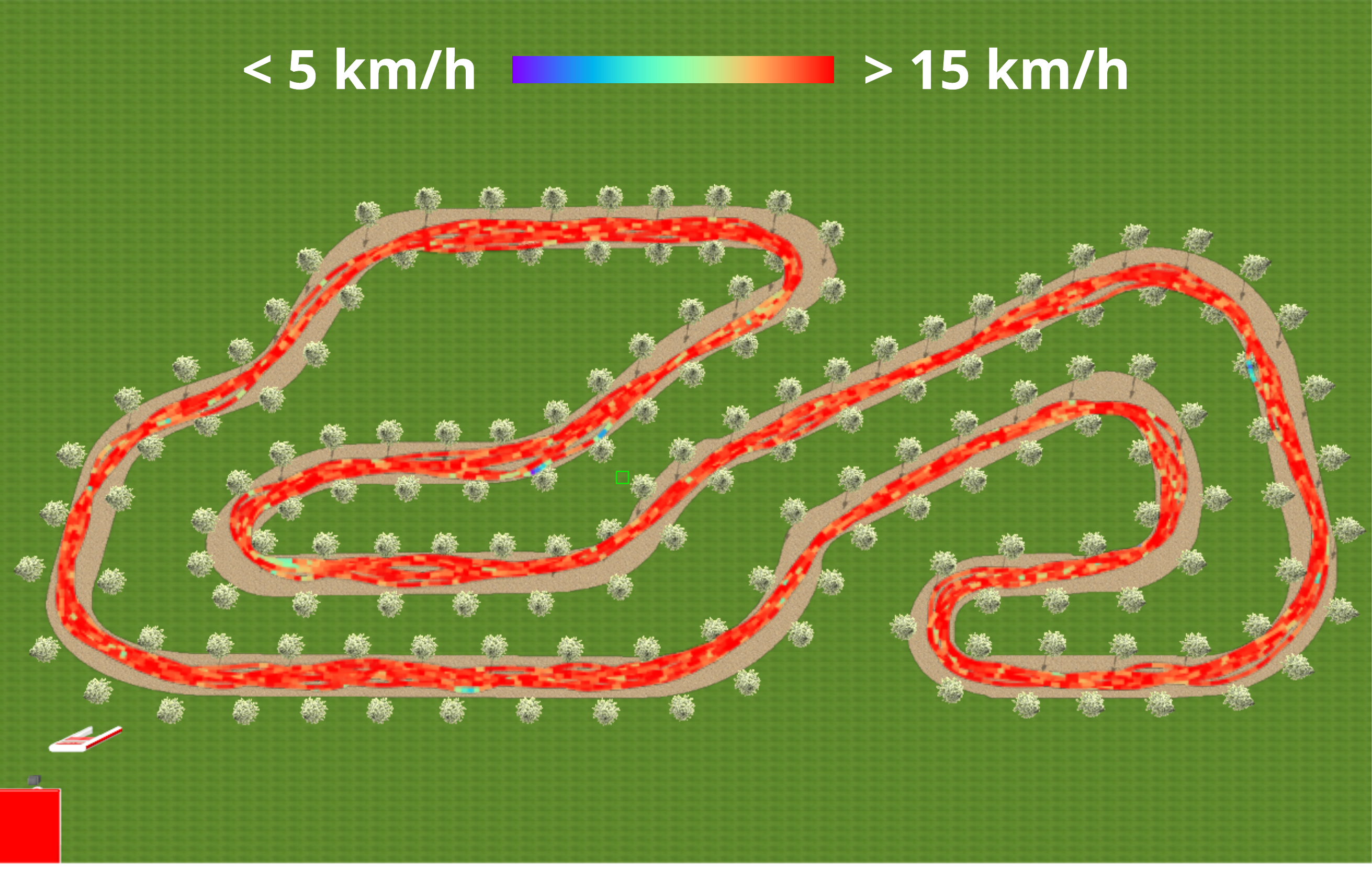}\label{fig:gazebo_traj}} \hfill
\subfloat[]{\includegraphics[width=0.34\textwidth]{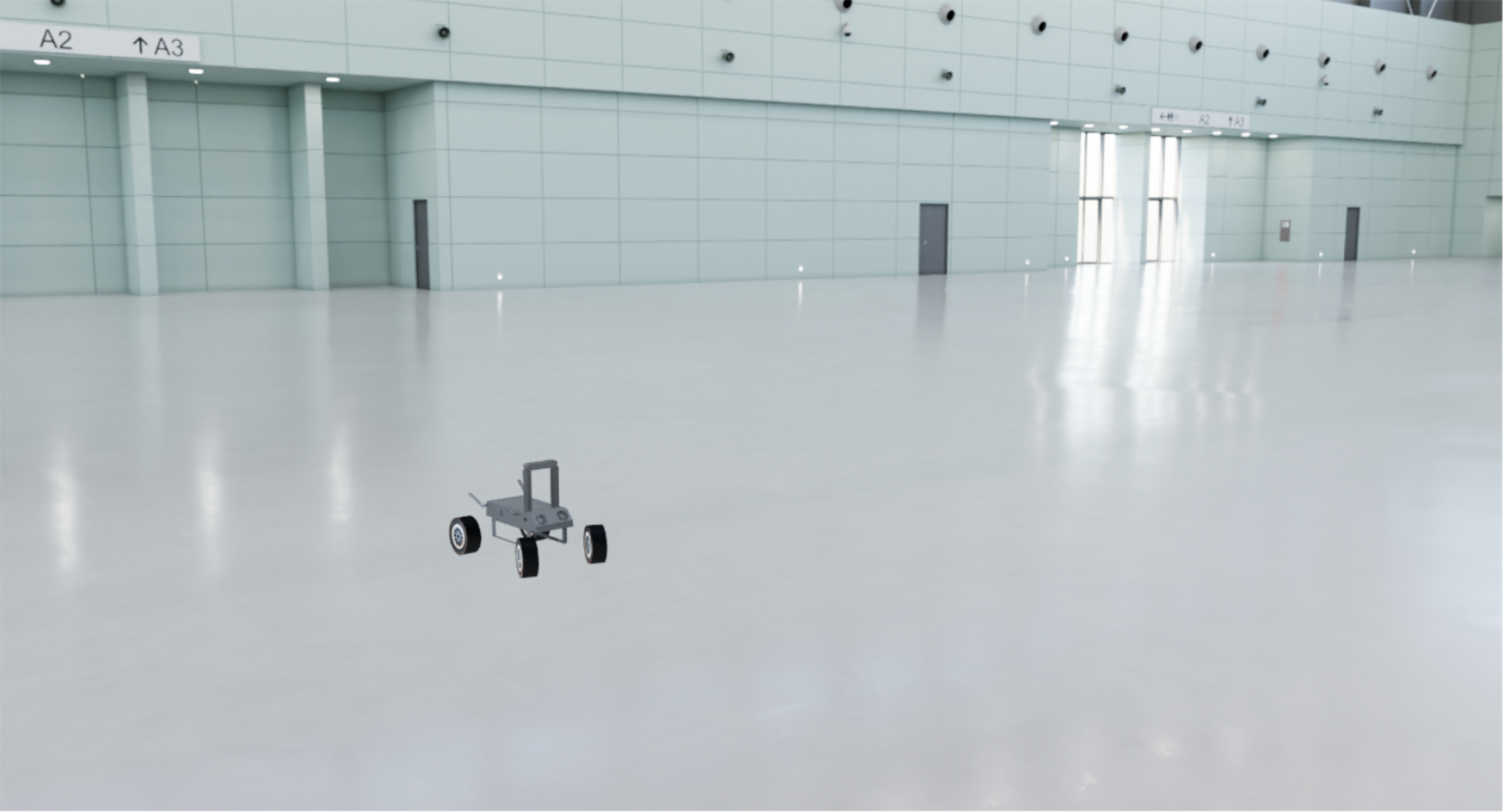}\label{fig:isaac_explore}}
\subfloat[]{\includegraphics[width=0.34\textwidth]{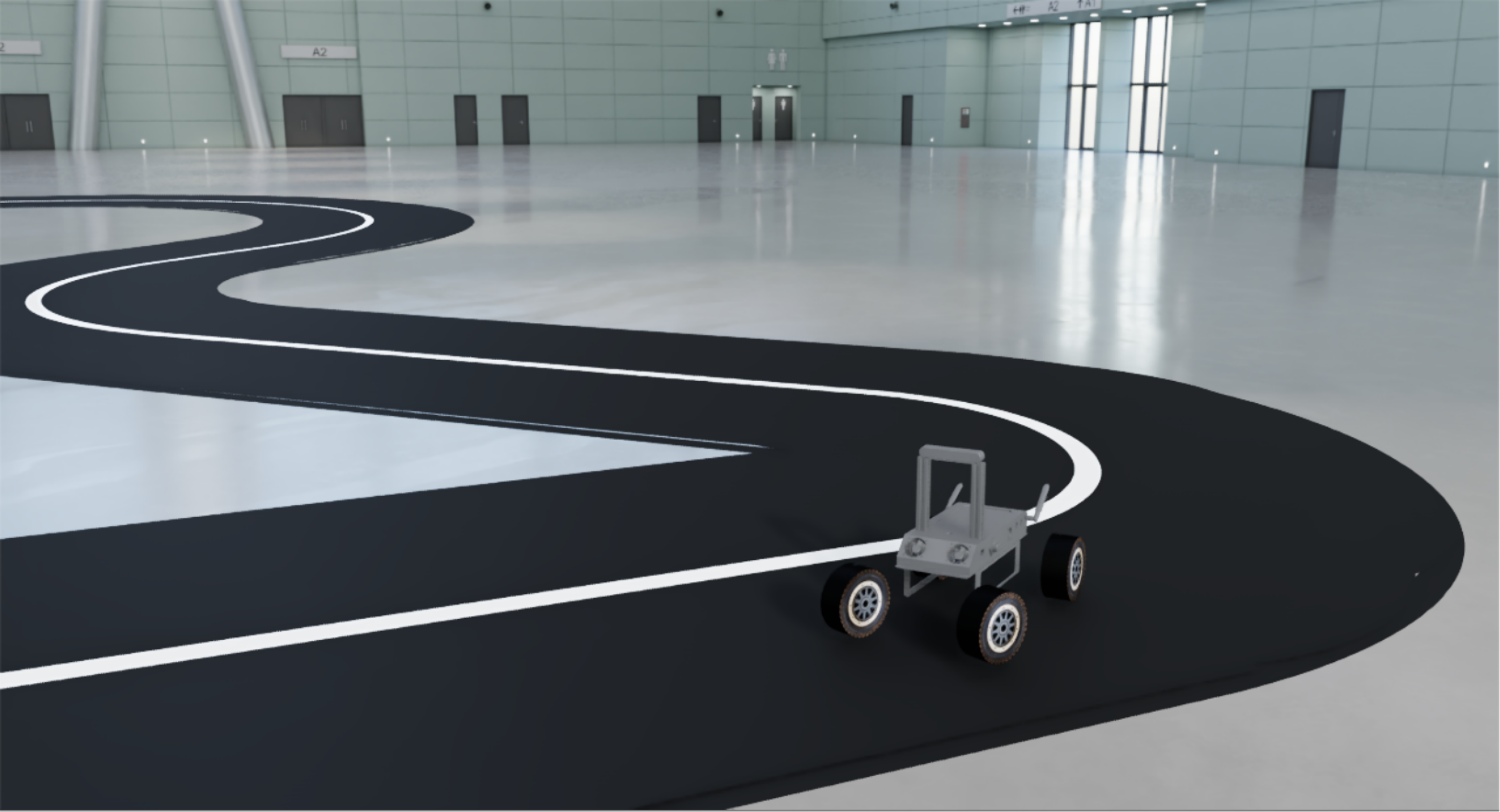}\label{fig:isaac_deploy}} 
\subfloat[]{\includegraphics[width=0.29\textwidth]{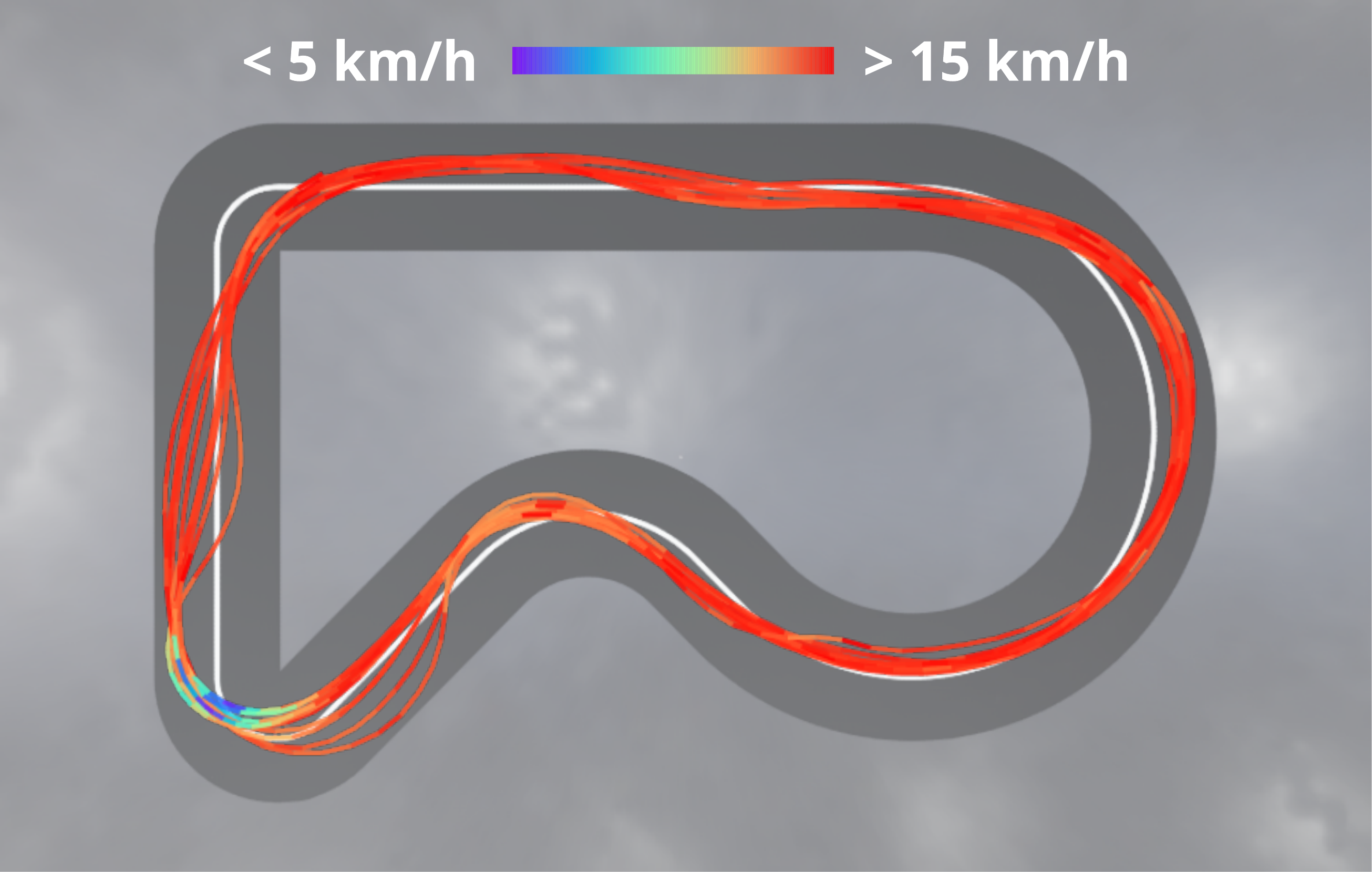}\label{fig:isaac_traj}}
\caption{
    Additional experiments using a 1:5 scale wheeled robotics testbed in (a)-(c) Gazebo and (d)-(f) Nvidia Isaac Sim.
    (a),(d) The simulated environments during the exploration phase. (b),(e) The simulated environments during the deployment phase. (c),(f) The vehicle trajectories taken by the robot at 300~iterations for driving 10~laps in a counter-clockwise direction.
}
\label{fig:wheeled}
\end{figure*}

\textbf{Experimental Results.}
We visualize the trajectories taken by the compared methods in Fig.~\ref{fig:safe}. To illustrate the vehicle's stability, we display the rotational impacts exerted on the vehicle during driving onto the trajectories. We simply use angular momentum, obtained by the squared sum of roll rate and pitch rate. We normalize the angular momentum by the maximum angular momentum value across all trials. Therefore, a trajectory with a darker color represents that the vehicle is more unstable. The angular and vertical motions of the vehicle during experiments are shown in Table~\ref{table:safe} for quantitative evaluation. 

Although UN can control the vehicle at the reference speed while avoiding collisions with obstacles and staying on track, it fails to circumvent uncertain regions because it is blind to uncertainty. It drives over the bumps at Scenario \#1 and Scenario \#5, eventually resulting in overturning. NP has opted for terrain with less uncertainty and, as a result, has chosen moderate terrain rather than the high bumps (see Scenario \#1 and Scenario \#5). However, it fails to circumvent the largest bump at Scenario \#3. On the other hand, Hybrid decides to slow down before the bump and take a detour to satisfy the safety criteria, despite the speed penalty. Conservative is a reasonable option in practical applications because it minimizes the damage to the robot. It takes the most secure trajectory as can be seen at Scenario \#4 and Scenario \#5 due to the low uncertainty threshold. The trajectory taken by Conservative also shows that the vehicle rarely experienced terrain impacts except at the inevitable bump at Scenario \#2. Hybrid shows a good balance between safety and agility. It avoids regions with high degrees of uncertainty, which might irreversibly damage or destroy the vehicle, while navigating the terrains with low levels of uncertainty at the desired speed. Table~\ref{table:safe} demonstrates that the methods with hard uncertainty penalty show a significantly lower degree of undesirable vertical motions and angular motions compared to the other methods.

\subsection{Experiments on Wheeled Robots \label{sec:exp_wheeled}}

In this section, we conduct a sanity check to ensure that our framework is scalable to other types of robot platforms. We use a 1:5 scale wheeled robotics testbed as the novel target system. We examine our framework in two robotics simulators: Gazebo~\cite{koenig_design_2004} and Nvidia Isaac Sim~\cite{makoviychuk_isaac_2021}. In each simulator, we conduct the active exploration experiments~\textbf{(Q1)} and direct deployment experiments~\textbf{(Q2)} with the best performing exploration strategy. Note that we do not modify any settings in our framework except for adjusting the target speed according to the robot's specifications~($v_{\text{target}}$ = 15~km/h) in both exploration and deployment phases. Other experimental settings are identical to those described in previous sections, including parameters of neural networks, controllers, and objective functions. Fig.~\ref{fig:wheeled} shows the simulated environments and the resulting trajectories in deployment tests taken by the robots. The results suggest that our framework has the potential to be applied to other robot platforms with continuous state and action space, actively exploring training data, and safely controlling the robot.

\section{Conclusion}

In this paper, we presented a unified model-based reinforcement learning framework with dynamics learning that bridges active exploration and uncertainty-aware deployment in the robotics control domain. The dynamics is learned using a fully parallelized probabilistic ensemble neural network that is sensitive to uncertainty. For active exploration, the epistemic uncertainty can be quantified by measuring the ensemble disagreement via Jensen-Rényi Divergence. Two opposing task objectives for exploration and deployment can be optimized by the state-of-the-art sampling-based MPC. Our extensive experiments demonstrate that the unified framework can be applied to both autonomous vehicles and wheeled robots. Our framework shows promising results for both exploration and deployment, in that the robot efficiently collects useful training data samples that are essential to achieving the task, and successfully avoids uncertain regions by imposing an uncertainty penalty. We hope that this work will serve as a stepping stone towards more general learning-based robotic applications under uncertainty.

\section{Appendix}

\subsection{Additional Details for Learned Dynamics \label{sec:appendix}}
To perform an ablation study on the neural network dynamics, we collect a human-controlled driving dataset. We build a race track in the IPG CarMaker simulator, modeled after a kart circuit located in Kirchlengern, Germany \cite{kim_smooth_2022}. The length of the track is 1016~m and it has two moderate curves and four sharp curves. The overall structure of the track is similar to that in Fig.~\ref{fig:wheeled}c. The driving data collected on the track consists of 1) zig-zag driving at low speeds, 2) high-speed driving, and 3) sliding maneuvers, in both clockwise and counter-clockwise directions. We collected a total of 70 minutes of data at a rate of 10~Hz. We split the data into 70$\%$ for training and 30$\%$ for testing after randomizing the data to remove temporal correlations.

We evaluate the test performance of the compared models with respect to each state variable in $\vx = (v_x, v_y, r)^{\top}$, which are the longitudinal velocity $v_x$, the lateral velocity $v_y$, and the yaw rate $r$. We use the Root Mean Square Error~(RMSE) as the performance metric. We also display the average test RMSE for the convenience of comparison.

\textbf{Model Comparison.}
First, we compare the pure MLP model and the Recurrent Neural Network~(RNN) model. In this experiment, we use a deterministic model without ensemble and use L2 loss function. The other parameters of the MLP model follow those mentioned in Section~\ref{sec:exp}. In RNN models, we replace the first layer of the MLP model as GRU~\cite{cho_properties_2014} and LSTM~\cite{hochreiter_long_1997}, respectively, with the same hidden layer size as the MLP model. The best test performances of each model during 1000 epochs are shown in Table~\ref{table:model}. The RNN models show slightly better performances than the MLP model. However, we choose MLP to benefit from the parallelizable structure of the ensemble MLP model.

\textbf{History Comparison.}
In this experiment, we use PENN model as described in Section~\ref{sec:method_implementing}. We evaluate the performances of the PENNs with different lengths of state-action history~$H$. The results are shown in Table~\ref{table:history}. The lengths of 3, 4, and 5 produce the best results among the compared models. When the history length exceeds 5, the prediction performance generally decreases. In accordance with prior literature employing the same strategy~\cite{spielberg_neural_2019, kim_toast_2022, spielberg_neural_2021, lambert_low-level_2019}, we determine the history length to be 4.

\begin{table}[t!]
\renewcommand\arraystretch{1.2}
\centering
\scriptsize
\caption{A comparison between the MLP model and the RNN models. The prediction performance on the test set is evaluated with the RMSE. }
\resizebox{0.8 \linewidth}{!}{
    \begin{tabular}{c|ccc}
        \toprule
        \textbf{Model} & MLP & GRU~\cite{cho_properties_2014} & LSTM~\cite{hochreiter_long_1997}\\
        \midrule
          Total &  0.0802 & 0.0721 & 0.0701 \\
           $v_x$ [m/s] &    0.0990 & 0.0861 & 0.0843 \\
           $v_y$ [m/s] &    0.06510 & 0.0587 & 0.0576\\
           $r$ [rad/s] &    0.0707 & 0.0660 & 0.0647 \\
        \bottomrule
    \end{tabular}
}
\label{table:model}
\end{table}

\begin{table*}[t!]
\renewcommand\arraystretch{1.2}
\centering
\scriptsize
\caption{A comparison between different lengths of state-action history $H$. The prediction performance on the test set is evaluated with the RMSE.}
\resizebox{ 0.95 \textwidth}{!}{
    \begin{tabular}{c|cccccccccc}
        \toprule
        \textbf{History Length} & 1 & 2 & \underline{\textbf{3}} & \underline{\textbf{4}} & \underline{\textbf{5}} & 6 & 7 & 8 & 9 & 10 \\
        \midrule
          Total &  0.0623 & 0.0589 & \textbf{0.0423} & 0.0453 & 0.0437 & 0.0667 & 0.0852 & 0.0528 & 0.0523 & 0.0511 \\
                               $v_x$ [m/s] &    0.0779 & 0.0662 & \textbf{0.0548} & 0.0610 & 0.0593 & 0.0826 & 0.0982 & 0.0734 & 0.0742 & 0.0709 \\
                               $v_y$ [m/s] &    0.0637 & 0.0691 & \textbf{0.0373 }& 0.0389 &\textbf{ 0.0373} & 0.0701 & 0.1024 & 0.0429 & 0.0408 & 0.0419 \\
                               $r$ [rad/s] &     0.0389 & 0.0357 & 0.0319 & 0.0305 & \textbf{0.0284} & 0.0399 & 0.0404 & 0.0336 & 0.0321 & 0.0327 \\
        \bottomrule
    \end{tabular}
    }
\label{table:history}
\end{table*}

\subsection{Sim-to-Real Transfer}
We successfully transferred our algorithm to real-world settings for uncertainty-aware deployment tasks. We integrated our algorithm with global path planning and online traversability map generation using a LiDAR sensor. These experiments were conducted on our off-road testbeds. The experimental results can be found on our project page: \url{https://taekyung.me/rss2023-bridging}.

\section{Acknowledgment}
This work was supported by the Korean Government~(2023).

\addtolength{\textheight}{0 cm}   % This command serves to balance the column lengths

%%%%%%%%%%%%%%%%%%%%%%%%%%%%%%%%%%%%%%%%%%%%%%%%%%%%%%%%%%%%%%%%%%%%%%%%%%%%%%%%

%%%%%%%%%%%%%%%%%%%%%%%%%%%%%%%%%%%%%%%%%%%%%%%%%%%%%%%%%%%%%%%%%%%%%%%%%%%%%%%%

%%%%%%%%%%%%%%%%%%%%%%%%%%%%%%%%%%%%%%%%%%%%%%%%%%%%%%%%%%%%%%%%%%%%%%%%%%%%%%%%

% \section*{ACKNOWLEDGMENT}
% This work was supported by Agency for Defense Development.

% disable magenta url color for reference
\hypersetup{
    urlcolor = .,
}

%%%%%%%%%%%%%%%%%%%%%%%%%%%%%%%%%%%%%%%%%%%%%%%%%%%%%%%%%%%%%%%%%%%%%%%%%%%%%%%%
%\bibliographystyle{plainnat}
\bibliographystyle{IEEEtran}
%\typeout{}
\bibliography{mybib_wo_link.bib}
%\bibliography{references.bib}

\end{document}